\documentclass[letterpaper]{article} 
\usepackage{aaai2026}  
\usepackage{times}  
\usepackage{helvet}  
\usepackage{courier}  
\usepackage[hyphens]{url}  
\usepackage{graphicx} 
\usepackage{natbib}  
\usepackage{caption} 
\usepackage{algorithm}
\usepackage{algorithmic}

\usepackage{newfloat}
\usepackage{listings}

\usepackage{amsmath}
\usepackage{multirow}
\usepackage{booktabs}
\usepackage{url}
\usepackage{subcaption}
\usepackage[misc]{ifsym}
\usepackage{amssymb}

\DeclareCaptionStyle{ruled}{labelfont=normalfont,labelsep=colon,strut=off} 
\floatstyle{ruled}
\newfloat{listing}{tb}{lst}{}
\floatname{listing}{Listing}
\begin{document}
%
\title{Think-J: Learning to Think for Generative LLM-as-a-Judge}


\author{
    Hui Huang\textsuperscript{1}$^{*}$, 
    Yancheng He\textsuperscript{1}$^{*}$, 
    Hongli Zhou\textsuperscript{1}$^{*}$,
    Rui Zhang\textsuperscript{1},
    Wei Liu\textsuperscript{1}, \\
    Weixun Wang\textsuperscript{1},
    Jiaheng Liu\textsuperscript{2$\dagger$},
    Wenbo Su\textsuperscript{1}
}

\affiliations{
    \textsuperscript{\rm 1}Alibaba Group, Hangzhou, China \\
    \textsuperscript{\rm 2}School of Intelligence Science and Technology, Nanjing University, Suzhou, China \\
    \text{hh456524@taobao.com, liujiaheng@nju.edu.cn}

}

\maketitle

\let\oldthefootnote\thefootnote
\let\thefootnote\relax\footnotetext{$*$ Equal contribution.}
\let\thefootnote\relax\footnotetext{$\dagger$\:Corresponding Author.}
\let\thefootnote\oldthefootnote

\begin{abstract}
LLM-as-a-Judge refers to the automatic modeling of preferences for responses generated by Large Language Models (LLMs), which is of significant importance for both LLM evaluation and reward modeling. Although generative LLMs have made substantial progress in various tasks, their performance as LLM-Judge still falls short of expectations. 
In this work, we propose Think-J, which improves generative LLM-as-a-Judge by learning how to think. We first utilized a small amount of curated data to develop the model with initial judgment thinking capabilities. Subsequently, we optimize the judgment thinking traces based on reinforcement learning (RL). We propose two methods for judgment thinking optimization, based on offline and online RL, respectively. The offline method requires training a critic model to construct positive and negative examples for learning. The online method defines rule-based reward as feedback for optimization.
Experimental results showed that our approach can significantly enhance the evaluation capability of generative LLM-Judge, surpassing both generative and classifier-based LLM-Judge without requiring extra human annotations.
\end{abstract}

\begin{links}
    \link{Code}{https://github.com/huihuichyan/think-j}
\end{links}

\section{Introduction}
As the capabilities of generative LLMs continue to advance, accurately evaluating the response quality has emerged as a crucial challenge\cite{huang-etal-2025-empirical}. This is not only vital for more efficient model development and comparison but also essential in the context of Reinforcement Learning from Human Feedback (RLHF), which relies on precise preference modeling as guidance \cite{wang2024secretsrlhflargelanguage,he2025can,liu2025itrickstrapsdeep}. However, traditional evaluation methods for generative models, such as BLEU \cite{papineni-etal-2002-bleu}, are based on predefined reference answers, which are often unavailable in open-ended scenarios.

\begin{figure}[t]
    \centering
        \includegraphics[width=0.92\linewidth]{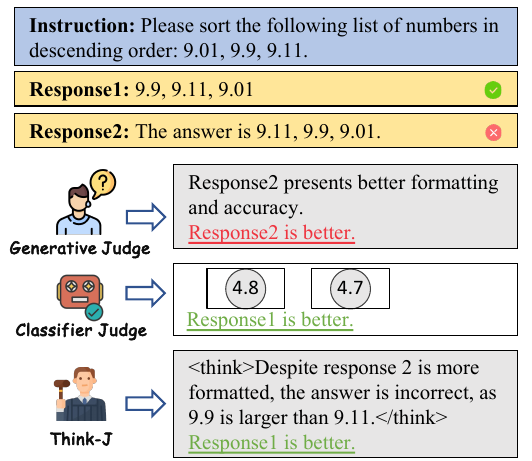}
        \caption{Comparison of different judge models. Our proposed Think-J takes into account both accuracy and interpretability based on thinking optimization.}
        \vspace{-3mm}
    \label{figure:illustration}
\end{figure}

Some studies have proposed LLM-as-a-Judge \cite{zheng2023judging}, which leverages the generative capabilities of LLMs for evaluating response quality. These work either directly leverage proprietary LLMs or fine-tune a smaller judge based on preference data \cite{gu2024survey}. However, the accuracy of their judgments remains unsatisfactory as revealed by recent benchmarks \cite{lambert2024rewardbench}. Other research has suggested fine-tuning a classifier based on preference data \cite{liu2024skywork}. While this method can achieve higher judgment accuracy, it lacks interpretability due to its scalar output, and the performance is highly dependent on the data quality \cite{wang-etal-2024-reward-modeling}.

Inspired by recent reasoning models such as o1 \cite{jaech2024openai} and Deepseek-R1 \cite{guo2025deepseek}, in this work, we propose Thinking-enhanced Generative Judge (Think-J), as shown in Figure \ref{figure:illustration}. Think-J aims to train a better generative judge by optimizing the model's judgment thinking capabilities. Specifically, Think-J consists of two steps: 

\textbf{1) Judgment Thinking Initialization}. We carefully curated 707 samples from preference data, considering various aspects such as accuracy, difficulty, and diversity. After that, thinking trace is annotated by proprietary models to initialize the thinking capabilities of the judge. 


\textbf{2) Judgment Thinking Optimization}. Due to the lack of high-quality critique annotation in preference datasets, we opt to optimize judgment thinking ability based on reinforcement learning (RL). Specifically, we adopted two methods: a) Critic-guided Offline Learning, leveraging an additional critic model to generate corresponding thinking traces based on provided judgment results, thus constructing positive and negative examples for offline RL. b) Rule-based Online Learning, defining rule-based rewards based on the correctness of judgment results, thus optimizing the thinking trace by online RL \cite{deepseek-math}.

We conducted experiments on three open-source models, and results showed that our proposed Think-J significantly outperformed existing LLM-judges with only limited training data. We also verify the effectiveness of our method compared with generative and classifier-based preference modeling methods. Our contributions are as follows:

\begin{enumerate}
    \item We propose to stimulate the judgment thinking ability of generative models with carefully curated data.
    \item We propose to optimize the judgment thinking ability of generative models with reinforcement learning.
    \item Our proposed Think-J significantly outperforms previous generative and classifier-based LLM-as-judges.
\end{enumerate}
\section{Background}

After the emergence of LLMs, numerous efforts have been made to design a more effective method for LLM evaluation \cite{chang2023surveyevaluationlargelanguage}. One of the most scalable and effective methods is LLM-as-a-Judge \cite{alpaca_eval,zheng2023judging}, namely utilizing proprietary LLMs, especially GPT4 \cite{achiam2023gpt}, to evaluate the LLM's response. For example, AlpacaEval \cite{alpaca_eval} used the win rate compared with baseline response determined by GPT-4 as the evaluation result. MT-Bench \cite{zheng2023judging} automatically scored the model's answers using GPT-4 as the results. The GPT-4-based evaluator is proven to presents comparable or even better consistency compared with human. 

However, relying on external API for evaluation may introduce consideration about privacy leakage, and the opacity of API models also challenges the evaluation reproducibility. Therefore, follow-up works suggest fine-tuning language models locally for evaluations, including JudgeLM \cite{zhu2023judgelm}, Auto-J \cite{li2023generative}, Prometheus \cite{kim2023prometheus}, Prometheus-2 \cite{zhu2023promptbench2}, OffsetBias \cite{park-etal-2024-offsetbias}, etc. These work typically construct preference data with judgment annotations and then finetune open-sourced LLMs to generate the judgment. Despite these fine-tuned judge models all achieve comparable accuracy with proprietary models, the evaluation is mostly conducted on the in-domain testsets, and these works are verified with a low scalability on more general benchmarks \cite{huang2024empiricalstudyllmasajudgellm}.

Another group of work fine-tunes a classifier on preference data based on the Bradley-Terry model, which is more commonly used on reward modeling. This approach is simple yet effective, as demonstrated on RewardBench \cite{lambert2024rewardbench} where most top-performing models are trained in a classification style \cite{liu2024skywork}. However, this method does not fully leverage the generative capabilities of LLMs, and is unable to provide rationales for its judgments, which is crucial for scalable evaluation. While recent work has begun to leverage the generative abilities of LLMs to combine critiques for scalar reward prediction \cite{ke-etal-2024-critiquellm,ye2024improvingrewardmodelssynthetic}, these critiques are often distilled from stronger proprietary models and hardly influence the final prediction \cite{liu2025inferencetimescalinggeneralistreward}. Effectively integrating the generative abilities of LLMs into evaluation remains an open challenge \cite{chen2025judgelrmlargereasoningmodels,whitehouse2025j1incentivizingthinkingllmasajudge,chen2025rmr1rewardmodelingreasoning,wang2025unifiedmultimodalchainofthoughtreward}.
\begin{figure*}[!t]
    \centering
        \includegraphics[width=0.95\linewidth]{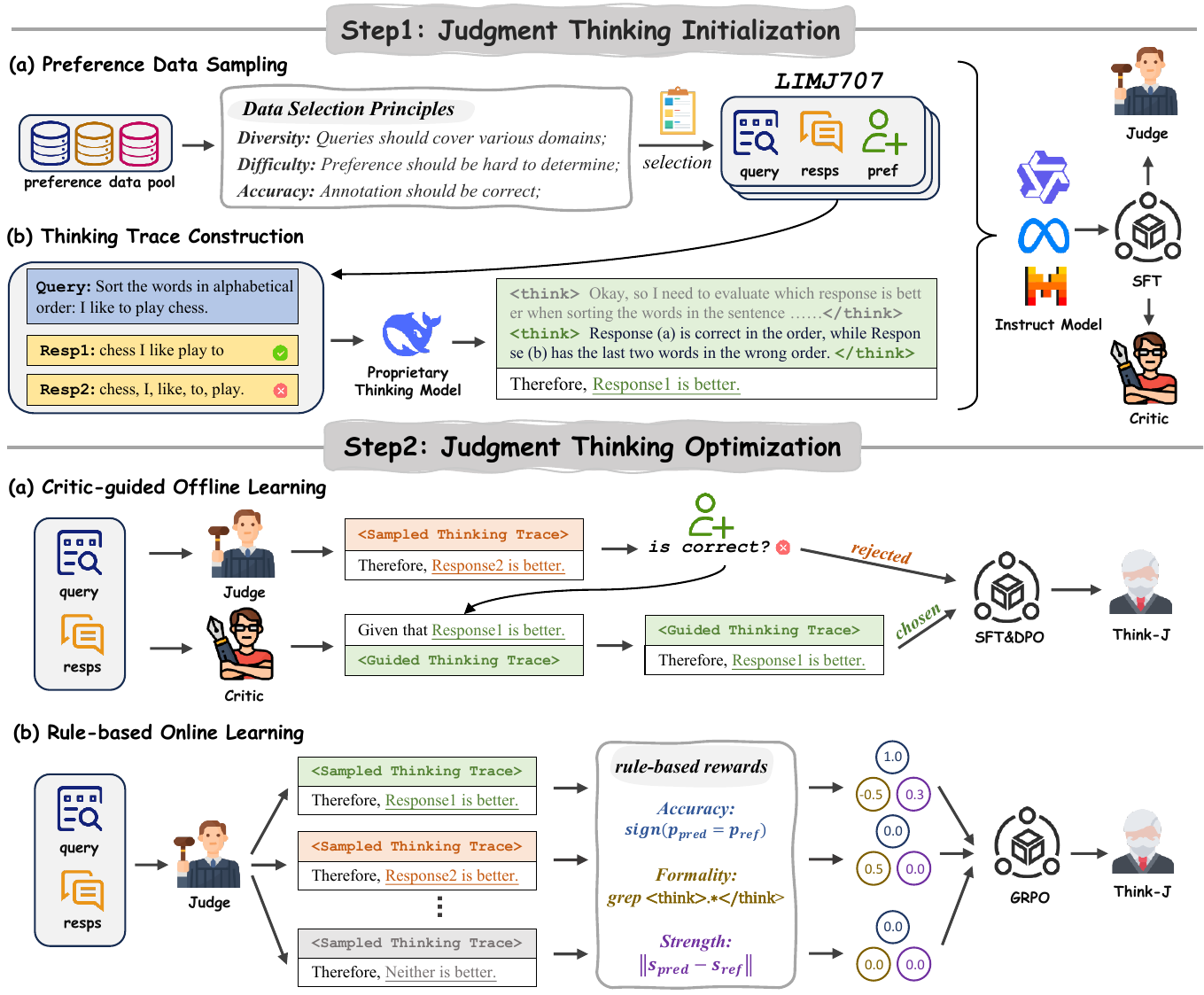}
        \caption{The illustration of our proposed framework. We begin by constructing high-quality judgment thinking traces using curated principles and proprietary thinking models. Based on this data, we initialize a judge model and a critic model, both equipped with judgment thinking capability. After that, we optimize the capability of the judge model through two methods: Critic-guided Offline Learning and Rule-based Online Learning, resulting in Think-J.}
        \vspace{-3mm}
    \label{figure:main-fig}
\end{figure*}

\section{Methodology}
\subsection{Judgment Thinking Initialization}
\label{sec:think-init}

Recent studies have shown that LLMs inherently possess long chain-of-thought (CoT) reasoning capabilities, which can be activated with a small amount of data \cite{muennighoff2025s1simpletesttimescaling,ye2025limoreasoning,liu2025air}. In this work, we also curate high-quality preference data, \textbf{LIMJ707}, to initialize the thinking capability of the judge model. Specifically, LIMJ707 is selected based on three principles:




\begin{itemize}
    \item \textbf{Accuracy}: The judgment (preference) annotation should be correct. We leverage the high-quality preference data Skywork-Preference-v0.2\footnote{\url{huggingface.co/datasets/Skywork/Skywork-Reward-Preference-80K-v0.2}}, which has been carefully validated to ensure accurate annotation.
    \item \textbf{Difficulty}: The sample should be sufficiently challenging. We apply the judge models to perform judgment for the sample three times, and select those samples where at least one judgment is failed, as these samples are likely more difficult and reflect the insufficiency of the judge.
    \item \textbf{Diversity}: The instruction should encompass various types to enhance judgment thinking capabilities in different aspects. We represent the instructions with an embedding model and then merge duplicate samples\footnote{For more details please refer to Appendix B.1.}.
\end{itemize}


The data statistics during processing are shown in Table \ref{tab:data-selection}. Based on LIMJ707, we construct judgment thinking trace by Deepseek-R1. After that, the annotated samples are used to initialize the model with judgment thinking capability by Supervised Fine-tuning \cite{ouyang2022training}\footnote{Due to the overly long thinking trace generated by Deepseek-R1, we performed trace-clipping to reduce training overhead and improve efficiency. Please refer to Appendix B.1 for more details.}.


\begin{table}[t]
\centering
\resizebox{0.45\textwidth}{!}{%
\begin{tabular}{@{\hspace{1em}}l@{\hspace{3em}}l@{\hspace{1em}}}
\toprule
Skywork-Reward-Preference-v0.2  & Data Num  \\ \hline
\: - initially selected         & 10K  \\ 
\: - filtered by difficulty     & 1496 \\
\: - filtered by diversity      & 870  \\
\: - filtered by accuracy       & 707  \\ \bottomrule
\end{tabular}}
\caption{Data statistics during constructing LIMJ707.}
\label{tab:data-selection}
\end{table}
\vspace{-2mm}


\subsection{Judgment Thinking Optimization}

To further enhance the alignment between the judge and human preference, the initialized thinking trace should be further optimized on preference data. However, preference data typically only includes binary labels without thinking trace annotations. Therefore, we propose judgment thinking optimization based on reinforcement learning (RL). Specifically, we propose two methods based on offline and online learning respectively, as shown in Figure \ref{figure:main-fig}.

\subsubsection{Critic-guided Offline Learning}

Offline RL methods represented by Direct Preference Optimization (DPO) \cite{rafailov2023direct} has been widely applied to LLM pipelines due to their efficiency and simplicity \cite{grattafiori2024llama}. In this work, we also aim to optimize the judgment thinking ability based on offline learning.

Due to the lack of golden thinking annotation in preference dataset\footnote{The rationale for fine-tuning an additional critic model over sampling to generate preference data is presented in Appendix A.3.}, we propose to train an additional critic model\footnote{Notice the critic model here differs from the critic model in traditional RL which is used for advantage estimation.}, to help to construct training samples. Both the critic and the judge models are trained on the same data (i.e., LIMJ707), with the following distinctions:



\begin{itemize}
    \item \textbf{Judge Model:} Given instruction-responses, it generates the thinking trace and judgment result.
    \item \textbf{Critic Model:} Given instruction-responses and judgment result, it generates the thinking trace.
\end{itemize}
    
Based on the two models, we can perform thinking optimization with the following steps:

\begin{enumerate}
    \item First, leverage the judge to evaluate the input to generate the thinking traces and results.
    \item If the result is correct, use the critic to generate an incorrect trace as the negative sample. Conversely, if the result is incorrect, use the critic to generate a correct trace as the positive sample. 
    \item Based on the positive and negative samples, optimize the judgment thinking ability with offline learning objective.
    \item These steps can be iterated to continuously enhance the judgment thinking capability.
\end{enumerate}

We adopt a combination of SFT and DPO as our training objective in this step:

\vspace{-3mm}
\noindent
\begin{small}
\begin{align}
& \mathcal{L}_\text{offline}(\pi_\theta; \mathcal{D}) = \nonumber \\
&\:\: - \mathbb{E}_{x \sim \mathcal{D}, (y_w, y_l) \sim \pi_{\theta} (y|x)} \left[ \log \pi_\theta(y_w \mid x) + \right. \nonumber \\
&\:\: \left. \log \sigma \left( \beta \log \frac{\pi_\theta(y_w \mid x)}{\pi_{ref}(y_w \mid x)} - \beta \log \frac{\pi_\theta(y_l \mid x)}{\pi_{ref}(y_l \mid x)} \right) \right]
\end{align}
\end{small}
\vspace{-2mm}

\noindent where $\pi_\theta$ and $\pi_{ref}$ represent the policy model and the reference model, and $y_w$ and $y_l$ denotes the positive and negative judgment traces, respectively.

With the help of critic model, samples with thinking annotation are constructed based on the correctness of judgment result. Therefore, the judge model will be enhanced to generate more accurate thinking for better judgment.


\subsubsection{Rule-based Online Learning}
\label{sec:online}
The recent success of R1-style methods have demonstrated the effectiveness of online RL using discrete, rule-based rewards \cite{deepseek-math}. In this work, we also apply online rule-based RL approach to optimize the judgment thinking capability. More specifically, we mainly utilize the GRPO algorithm, with the optimization objective as follows:

\vspace{-3mm}
\noindent
\begin{small}
\begin{align}
& J_{\text{online}}(\pi_\theta; \mathcal{D}) = \nonumber \\
&\:\: \mathbb{E}_{x \sim \mathcal{D},\{y_i\}_{i=1}^{G} \sim \pi_{\theta_{\text{old}}} (y|x)} \left[ \frac{1}{G} \sum_{i=1}^{G} \min \left( \frac{\pi_{\theta}(y_i|x)}{\pi_{\theta_{\text{old}}} (y_i|x)} A_i, \right. \right. \nonumber \\
&\:\: \left. \left. \text{clip} \left( \frac{\pi_{\theta}(y_i|x)}{\pi_{\theta_{\text{old}}} (y_i|x)}, 1 - \epsilon, 1 + \epsilon \right) A_i \right) - \beta D_{\text{KL}} (\pi_{\theta} || \pi_{\text{ref}}) \right]
\end{align}
\end{small}
\vspace{-1mm}

\noindent
where $G$ is group size, and $A_i$ is advantage. The reward function is designed as follows\footnote{We do not include a length penalty in rewards to encourage longer thinking, as we have observed that longer thinking does not necessarily lead to better accuracy in our case.}:

\vspace{-3mm}
\begin{small}
\begin{align}
r_{\text{accuracy}} &= \begin{cases}
    1, & \text{if } \text{judgement} = \text{label}\\
    0, & \text{if } \text{judgement} \neq \text{label}
\end{cases} \\
r_{\text{format}} &= \begin{cases}
    0, & \text{if } \text{format is right}\\
    -0.5, & \text{if } \text{format is wrong}
\end{cases} \\
r_\text{strength} &= ||\text{s}_{\text{pred}} - \text{s}_{\text{golden}}||, s \in \{1,2,3\}\\
r_{\text{final}} &= \alpha \cdot r_{\text{accuracy}} + \beta \cdot r_{\text{format}} + \gamma \cdot r_{\text{strength}}
\label{equation:reward}
\end{align}
\end{small}
\vspace{-3mm}

\noindent where detailed weights for different rewards are presented in Section 5.2. Notice we incorporate a reward for assessing preference strength, which is defined as the degree to which the judge favors one response over another\footnote{For example, a strength of 1 means the chosen response is only slightly better than the rejected, while a strength of 3 means the chosen is much better than the rejected. For more details about the criteria for the strength annotation, please refer to Appendix B.4.}. The prompt template of judgment is also adjusted as:

\vspace{1mm}
\texttt{<think> \{thinking trace\} </think>}

\texttt{Therefore, Response (a) is better, \underline{and} \underline{the preference strength is [[2]]}.}
\vspace{1mm}

Preference strength helps to perceive the relative quality of response pairs, without uniformly providing the same reward for different pairs despite the quality gap. We employ a comparatively simple scale of reward scores of reward, as during actual training, we observed that the model tends to manipulate scores towards extreme values\footnote{Please refer to Appendix A.1 for more details .}.

While absolute score annotation for a given response is challenging, annotating relative preference strength is more readily accessible \cite{wang2024helpsteer2,wang2025helpsteer3}. Moreover, if absolute score annotation is absent in the preference data, we can firstly fine-tune a BT-classifier based on the data, then leverage the classifier to assess the relative strength between chosen and rejected responses \cite{wang-etal-2024-reward-modeling}.
\begin{table*}[!htb]
\resizebox{1.0\textwidth}{!}{
\begin{tabular}{cc|ccccc|c|c}
\hline
\multirow{2}{*}{\textbf{Model}}                                                         & \multirow{2}{*}{\textbf{Sample Num}} & \multicolumn{5}{c|}{\textbf{RewardBench}}                                           & \textbf{RMBench} & \textbf{Auto-J}    \\
                                                                                        &                                      & \textbf{Chat} & \textbf{Hard} & \textbf{Safety} & \textbf{Reason} & \textbf{Overall} & \textbf{Overall}  & \textbf{Agreement} \\ \hline
Claude-3-5-Sonnet-20240620      & —                                    & 96.4          & 74.0          & 81.6            & 84.7            & 84.2            & 68.9             & 70.7               \\
Qwen-2.5-32B-Instruct                                                                  & —                                    & 96.2          & 74.0          & 88.7            & 86.9            & 86.5            & 68.3             & 59.6               \\
GPT-4o-2024-08-06               & —                                    & 96.1          & 76.1          & 88.1            & 86.6            & 86.7            & 68.8             & 69.8               \\
Gemini-1.5-Pro-0514             & —                                    & 92.3          & 80.6          & 87.9            & 92.0            & 88.2            & 74.4             & 68.1               \\ \hline
JudgeLM-33B \cite{zhu2023judgelm}                                      & 100K                                 & 90.1          & 51.0          & 85.7          & 39.7            & 66.6            & 49.6             & 45.3               \\
Prometheus-7b-v2.0 \cite{zhu2023promptbench2}                          & 40K                                  & 83.9          & 49.2          & 72.8          & 72.0            & 69.5            & 52.4             & 63.1               \\
Prometheus-8x7b-v2.0 \cite{zhu2023promptbench2}                        & 40K                                  & 93.0          & 47.1          & 80.5          & 77.4            & 74.5            & 57.4             & 68.5               \\
Llama-3-OffsetBias-8B \cite{park-etal-2024-offsetbias}                 & 276K                                 & 92.5          & 80.3          & 86.8          & 76.4            & 84.0            & 66.0             & 68.7               \\
CompassJudger-32B \cite{cao2024compass}                                & 2041K                                & 97.4          & 65.6          & 85.1          & 87.1            & 83.8            & 69.4             & \textbf{80.7}      \\
STE-Llama3.1-70B \cite{wang2024selftaughtevaluators}                   & 20K                                  & 96.9          & 85.1          & 89.6          & 88.4            & 90.0            & 65.3             & 72.0               \\
SynRM-Command-R-35B  \cite{ye2024improvingrewardmodelssynthetic}       & 5K                                   & 97.5          & 76.8          & 88.5          & 86.3            & 87.3            & —                & —                  \\
CLoud-Llama3-70B  \cite{ke-etal-2024-critiquellm}                      & 350K                                 & 98.0          & 75.6          & 87.6          & 89.0            & 87.6            & —                & —                  \\ \hline
Think-J-Qwen-2.5-32B (Helpsteer2-Pref)                                                 & 9.8K                                 & 96.7          & 83.2          & 90.1          & 92.0            & \textbf{90.5}   & \textbf{79.8}    & \textbf{75.8}      \\ \hline
\end{tabular}}
\caption{Experiment results of top LLM-Judges on RewardBench, RMBench, and Auto-J-test. In this table, we report the best performance achieved by various LLM-Judges, trained on different base models and datasets.}
\label{tab:sota-contrast}
\end{table*}

\begin{table*}[!htb]
\resizebox{1.0\textwidth}{!}{
\begin{tabular}{cc|ccccc|ccccc}
\hline
\multirow{3}{*}{\textbf{Model}}                                                 & \multirow{3}{*}{\textbf{Method}} & \multicolumn{5}{c|}{\textbf{HH-RLHF}}                                               & \multicolumn{5}{c}{\textbf{Helpsteer2-Pref}}                                  \\
                                                                                &                                  & \multicolumn{5}{c|}{\textbf{RewardBench}}                                           & \multicolumn{5}{c}{\textbf{RewardBench}}                                            \\
                                                                                &                                  & \textbf{Chat} & \textbf{Hard} & \textbf{Safety} & \textbf{Reason} & \textbf{Overall} & \textbf{Chat} & \textbf{Hard} & \textbf{Safety} & \textbf{Reason} & \textbf{Overall} \\ \hline
\multirow{9}{*}{\begin{tabular}[c]{@{}c@{}}Llama3-8B\\ -Instruct\end{tabular}}  & Direct Prompt (w/o CoT)          & 90.4          & 44.7          & 76.5            & 63.4            & 68.8            & 90.4          & 44.7          & 76.5            & 63.4            & 68.8            \\
                                                                                & Direct Prompt (w/ CoT)           & 84.6          & 40.9          & 50.8            & 58.8            & 58.8            & 84.6          & 40.9          & 50.8            & 58.8            & 58.8            \\ \cline{2-12} 
                                                                                & SFT on LIMJ707                   & 91.8          & 67.1          & 83.0            & 64.2            & 76.5            & 91.8          & 67.1          & 83.0            & 64.2            & 76.5            \\
                                                                                & SFT (w/o CoT)                    & 88.1          & 50.6          & 81.0            & 69.1            & 72.2            & 88.0          & 41.1          & 41.5            & 47.8            & 54.6            \\
                                                                                & SFT (w/ CoT)                     & 78.8          & 67.0          & 81.0            & 62.1            & 72.2            & 84.5          & 71.3          & 80.8            & 62.6            & 74.8            \\ \cline{2-12} 
                                                                                & BT Classifier                    & 81.8          & 72.6          & 79.3            & 85.2            & 79.7            & 88.6          & 73.9          & 81.2            & 91.8            & 83.9            \\
                                                                                & CLoud                            & 87.4          & 69.3          & 87.3            & 77.6            & 80.4            & 93.3          & 74.8          & 80.8            & 73.7            & 80.6            \\
                                                                                & SynRM                            & 87.2          & 74.3          & 81.1            & 80.2            & 80.7            & 91.9          & 68.4          & 80.4            & 87.9            & 82.2            \\ \cline{2-12} 
                                                                                & Think-J                          & 92.2          & 71.7          & 84.8            & 75.6            & \textbf{81.1}   & 93.9          & 74.3          & 90.3            & 77.8            & \textbf{84.1}   \\ \hline
\multirow{9}{*}{\begin{tabular}[c]{@{}c@{}}Qwen2.5-7B\\ -Instruct\end{tabular}} & Direct Prompt (w/o CoT)          & 94.4          & 56.9          & 81.0            & 77.3            & 77.4            & 94.4          & 56.9          & 81.0            & 77.3            & 77.4            \\
                                                                                & Direct Prompt (w/ CoT)           & 93.0          & 58.1          & 81.2            & 77.8            & 77.5            & 93.0          & 58.1          & 81.2            & 77.8            & 77.5            \\ \cline{2-12} 
                                                                                & SFT on LIMJ707                   & 89.5          & 75.2          & 80.7            & 74.5            & 80.0            & 89.5          & 75.2          & 80.7            & 74.5            & 80.0            \\
                                                                                & SFT (w/o CoT)                    & 89.8          & 74.3          & 82.5            & 61.3            & 77.0            & 95.0          & 75.9          & 87.2            & 75.8            & 83.5            \\
                                                                                & SFT (w/ CoT)                     & 94.1          & 74.0          & 86.0            & 64.7            & 79.7            & 88.7          & 69.7          & 84.5            & 76.1            & 79.8            \\ \cline{2-12} 
                                                                                & BT Classifier                    & 82.4          & 69.7          & 87.0            & 76.9            & 79.0            & 95.0          & 67.8          & 85.0            & 61.4            & 77.3            \\
                                                                                & CLoud                            & 85.2          & 79.2          & 86.6            & 57.9            & 77.2            & 94.7          & 73.3          & 82.3            & 59.0            & 77.3            \\
                                                                                & SynRM                            & 90.5          & 65.1          & 77.8            & 70.6            & 76.0            & 96.7          & 61.4          & 82.8            & 61.4            & 75.6            \\ \cline{2-12} 
                                                                                & Think-J                          & 94.4          & 70.2          & 83.8            & 79.8            & \textbf{82.0}   & 96.1          & 78.6          & 85.9            & 80.4            & \textbf{85.3}   \\ \hline
\end{tabular}}
\caption{Experiment results of different LLM judge training methods on RewardBench. In this table, we report the performance of various methods trained on the same base models (Llama3-8B-Instruct and Qwen2.5-7B-Instruct) and datasets (HH-RLHF and Helpsteer2-Pref) to enable a more direct comparison.}
\label{tab:main}
\end{table*}

\section{Experiments}
\subsection{Set-up}

We mainly conducted experiments on two popular open-sourced models with their instruction version: Qwen-2.5 \cite{qwen2025qwen25technicalreport} and Llama-3 \cite{grattafiori2024llama}. 

We primarily conducted our optimization on two datasets: HelpSteer2-Pref\footnote{\url{huggingface.co/datasets/nvidia/HelpSteer2}} and HH-RLHF\footnote{\url{huggingface.co/datasets/Anthropic/hh-rlhf}}. For a fair comparison, LIMJ707 was mixed into all training sets. While larger preference datasets are available, we excluded them from our training because our primary objective was to explore the most effective method for training LLM judges.



We mainly compare Think-J with the following generative LLM-judge approaches:

\begin{itemize}
    \item \textbf{Direct Prompt} Leverage the LLM to directly generate judgment without fine-tuning.
    \item \textbf{SFT (w/o CoT)} Train a generative model by supervised fine-tuning to perform judgment without thinking traces.
    \item \textbf{SFT (w/ CoT)} Train a generative model on correct judgment thinking traces and results.
\end{itemize}

We also compare Think-J with the following classifier-based LLM-judge approaches:

\begin{itemize}
    \item \textbf{BT Classifier} Feed the instruction and responses into the model and added a classification head, training it according to Bradley-Terry model \cite{sun2025rethinkingbradleyterrymodelspreferencebased}.
    \item \textbf{CLoud} \cite{ke-etal-2024-critiquellm} First train the model to generate critiques, and then leverage the critiques as additional input to improve the BT Classifier.
    \item \textbf{SynRM} \cite{ye2024improvingrewardmodelssynthetic} Leverage critiques generated by a proprietary model as additional input to improve the BT Classifier. We use Deepseek-R1 to generate the critiques.
\end{itemize}

To further showcase Think-J's evaluation capabilities, we also compared its 32B version against leading proprietary judge models, including closed-source options like Claude 3.5 Sonnet, GPT-4o, and Gemini 1.5 Pro, as well as open-source fine-tuned judges such as JudgeLM, Prometheus, and CompassJudger. These models are widely employed as LLM-as-a-Judge across diverse tasks.

We mainly perform evaluation on RewardBench \cite{lambert2024rewardbench}. We also evaluate our model on RMBench \cite{liu2025rmbench} and Auto-J-test \cite{li2023generative}. 

\textbf{We report the best results achieved by either offline or online learning for the main experiments by default.}

\subsection{Main Experiment}

\begin{table*}[!t]
\centering
\resizebox{0.95\textwidth}{!}{
\begin{tabular}{ccccccccccc}
\hline
\textbf{}                        & \multicolumn{5}{c}{\textbf{Llama-3-8B-Instruct}}                                     & \multicolumn{5}{c}{\textbf{Qwen-2.5-7B-Instruct}}                                    \\
\multirow{2}{*}{\textbf{Method}} & \multicolumn{5}{c}{\textbf{RewardBench}}                                             & \multicolumn{5}{c}{\textbf{RewardBench}}                                             \\
                                 & \textbf{Chat} & \textbf{Hard} & \textbf{Safety} & \textbf{Reason} & \textbf{Overall} & \textbf{Chat} & \textbf{Hard} & \textbf{Safety} & \textbf{Reason} & \textbf{Overall} \\ \hline
baseline                         & 90.4          & 44.7          & 76.5            & 63.4            & 68.8             & 94.4          & 56.9          & 81.0            & 77.3            & 77.4             \\ \hline
offline (Critic-guided)          & 95.0          & 73.3          & 87.4            & 77.2            & 83.2             & 94.8          & 74.2          & 83.5            & 80.5            & 83.3             \\
offline (w/o SFT)                & 95.1          & 63.8          & 88.3            & 77.6            & 81.2             & 95.7          & 73.8          & 86.0            & 74.6            & 82.5             \\ \hline
online (PPO)                     & 70.4          & 75.3          & 77.2            & 70.0            & 73.2             & 88.7          & 69.7          & 84.5            & 76.1            & 79.8             \\
online (Reinforce++)             & 93.7          & 74.3          & 89.7            & 77.9            & 84.0             & 94.7          & 70.6          & 90.4            & 82.3            & 84.5             \\
online (GRPO)                    & 93.9          & 74.3          & 90.3            & 77.8            & \textbf{84.1}    & 96.1          & 78.6          & 85.9            & 80.4            & \textbf{85.3}    \\ \hline
\end{tabular}}
\caption{Experiment results of different RL strategies.}
\label{tab:RL-compare}
\end{table*}

As demonstrated in Table \ref{tab:sota-contrast}, Think-J-32B achieved the best performance across all benchmarks, surpassing both close-sourced and fine-tuned judges. Notably, our method required only 9832 training samples, but still achieve marginal improvement on 32B-sized models. This underscores the effectiveness of Think-J, which leverages thinking optimization with the correctness of judgment as feedback to enhance preference modeling capability.

Furthermore, as Table \ref{tab:main} illustrates, starting from the same base model and training data, our proposed method consistently outperforms other approaches for fine-tuning LLM judges. This demonstrates the effectiveness of our judgment thinking optimization. In contrast, both naive and fine-tuned generative methods yield inferior results. Notably, the rejection sampling method, despite being trained on the same data constructed by the critic, also underperforms. This underscores the critical role of learning from negative samples when modeling human preference \cite{liu2024statistical}.

On the other hand, the classifier-based method achieves relatively higher results but can only produce numerical outputs that lack interpretability. Additionally, the reasoning-enhanced classifiers, including CLoud and SynRM, performs worse than expected. This suggests that combining generative CoT into classifier may introduce noise rather than useful information for classification.
\section{Analysis}
\subsection{Less is More for Thinking Initialization}

\begin{figure}[!t] 
    \centering 

    \begin{subfigure}[b]{0.46\textwidth}
        \centering
        \includegraphics[width=1.0\linewidth]{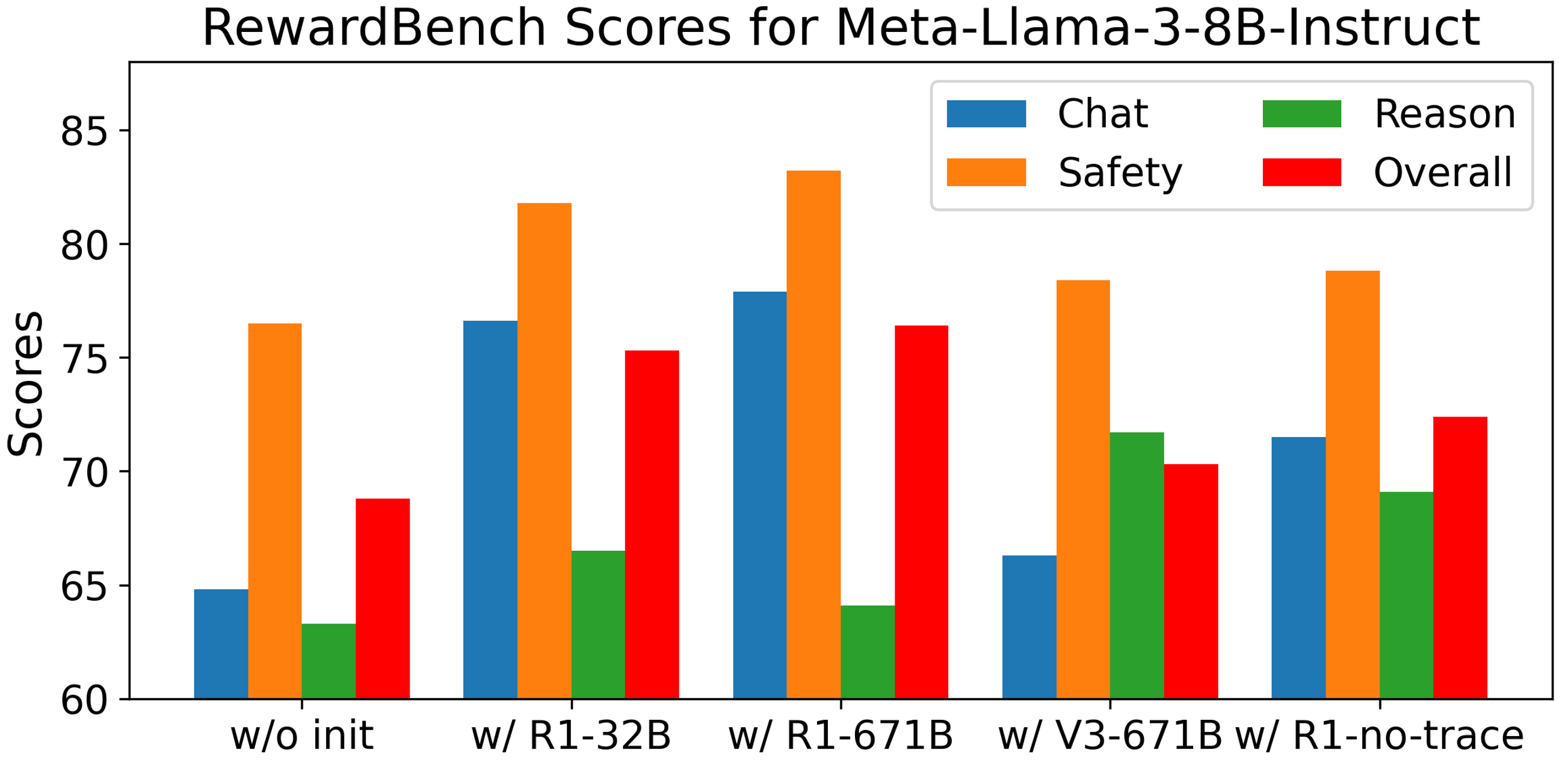}
    \end{subfigure}
    \hfill
    \begin{subfigure}[b]{0.46\textwidth}
        \centering
        \includegraphics[width=1.0\linewidth]{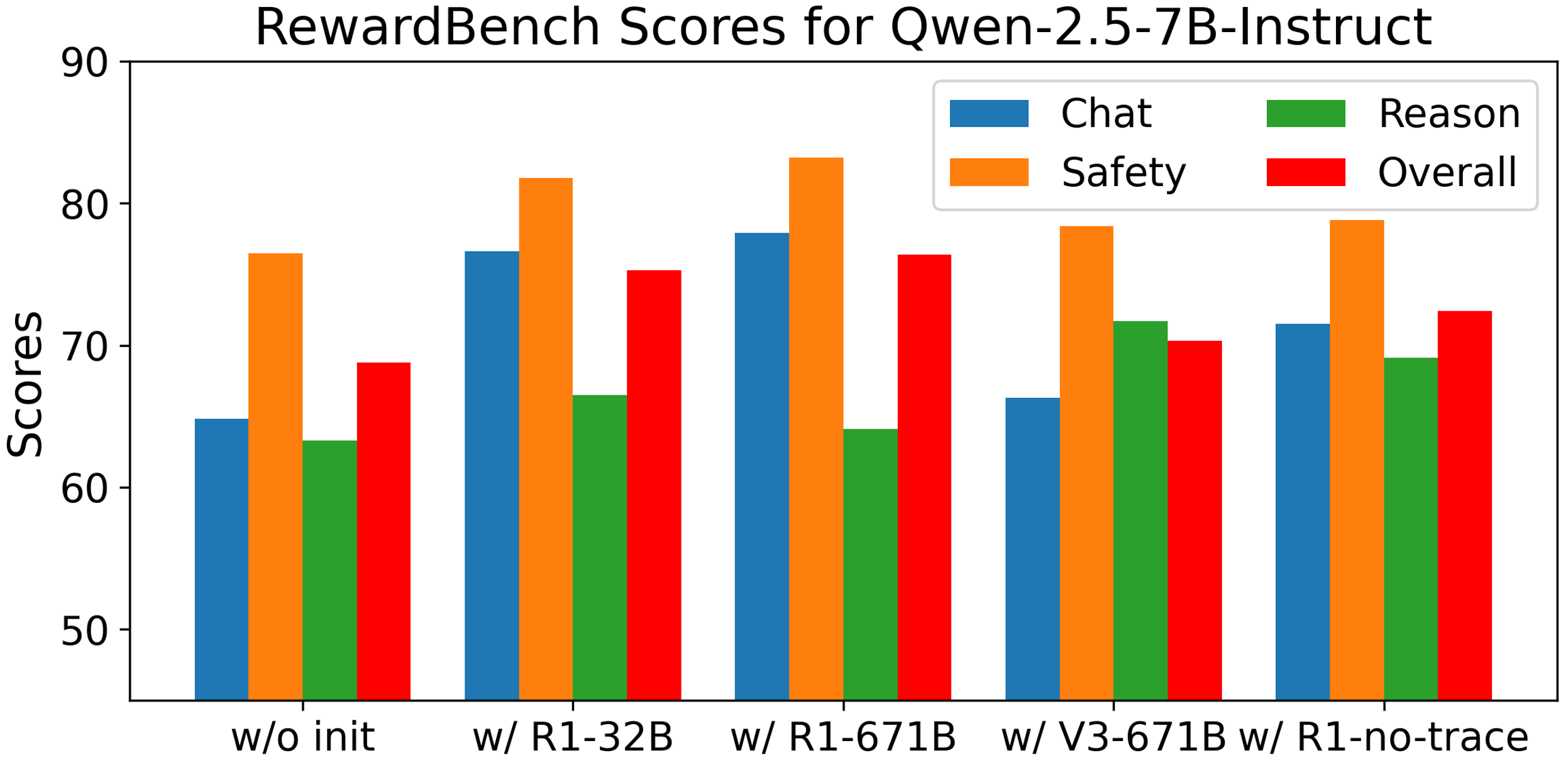}
    \end{subfigure}
    \caption{Experiment results of different data source for judgment thinking initialization.}
    \label{figure:init-compare}
\end{figure}

We compared the performance of different data source for judgment thinking initialization in Figure \ref{figure:init-compare}. Our findings indicate that a small amount of thinking trace annotated with Deepseek-R1 can significantly enhance the model's judgment capabilities. In contrast, using thinking trace annotated with Deepseek-V3, or removing the trace from the data would results in a substantial decline in performance. This highlights the importance of high-quality thinking trace for judgment thinking initialization. 


We also compare the impact of different data selection strategies in Table \ref{tab:selection}. The results show that data quality is crucial for effective model initialization. For instance, a model initialized with chatbot-arena\cite{chiang2024chatbotarenaopenplatform}, which contains relatively noisy data, achieves minimal improvement. Conversely, selecting the longest traces proves detrimental, as the longest traces are often code or math-related, which can negatively impact the data diversity.

\begin{table}[!tb]
\resizebox{0.47\textwidth}{!}{
\begin{tabular}{ccccc}
\hline
\multirow{2}{*}{\textbf{Method}} & \multicolumn{4}{c}{\textbf{RewardBench}}                         \\
                                & \textbf{Chat} & \textbf{Safety} & \textbf{Reason} & \textbf{Overall} \\ \hline
Llama-3-8B-Inst                 & 64.8          & 76.5            & 63.4            & 68.8         \\ \hline
\multicolumn{5}{l}{\textit{Init with 1000 samples on chatbot-arena}}                                               \\ \hline
random-sampled                  & 66.0          & 74.3            & 64.4            & 68.3         \\ \hline
\multicolumn{5}{l}{\textit{Init with 1000 samples on skywork-preference-v3}}                                                  \\ \hline
random-sampled                  & 75.1          & 82.0            & 66.4            & 75.4          \\
longest                         & 77.2          & 79.3            & 61.8            & 74.2          \\
Llama3-failed                   & 76.4          & 85.0            & 66.5            & \textbf{76.1} \\ \hline
\end{tabular}}
\caption{Experiment results of different data selection strategies for judgment thinking initialization.}
\label{tab:selection}
\end{table}

\begin{table}[!tb]
\resizebox{0.47\textwidth}{!}{
\begin{tabular}{cccccc}
\hline
\multirow{2}{*}{\textbf{Init}} & \multirow{2}{*}{\textbf{RL}} & \multicolumn{4}{c}{\textbf{RewardBench}}                    \\
                               &                              & \textbf{Chat} & \textbf{Safety} & \textbf{Reason} & \textbf{Overall} \\ \hline
\multicolumn{2}{c}{Llama-3-8B-Inst}                           & 64.8          & 76.5            & 63.4            & 68.8    \\ \hline
\multirow{2}{*}{no init}       & offline                      & 80.3          & 73.2            & 79.4            & 79.1    \\
                               & online                       & 80.8          & 86.8            & 71.8            & 80.8    \\ \hline
\multirow{2}{*}{w/ init}       & offline                      & 82.8          & 87.4            & 77.2            & 83.2    \\
                               & online                       & 82.9          & 90.3            & 77.8            & \textbf{84.1}   \\ \hline
\end{tabular}}
\caption{Experiment results of different initialization strategies for RL optimization on HelpSteer2-Pref.}
\label{tab:init4RL}
\end{table}

\begin{table*}[!ht]
\centering
\resizebox{0.85\textwidth}{!}{
\begin{tabular}{c|ccccc|ccccc}
\hline
\multirow{3}{*}{\textbf{Setting}} & \multicolumn{5}{c|}{\textbf{Qwen2.5-32B-Instruct}}                                   & \multicolumn{5}{c}{\textbf{Qwen2.5-7B-Instruct}}                                     \\
                                  & \multicolumn{5}{c|}{\textbf{RewardBench}}                                            & \multicolumn{5}{c}{\textbf{RewardBench}}                                             \\
                                  & \textbf{Chat} & \textbf{Hard} & \textbf{Safety} & \textbf{Reason} & \textbf{Overall} & \textbf{Chat} & \textbf{Hard} & \textbf{Safety} & \textbf{Reason} & \textbf{Overall} \\ \hline
$\beta$=1.0                       & 96.7          & 83.2          & 90.1            & 92.0            & \textbf{90.5}    & 96.1          & 78.6          & 85.9            & 80.4            & \textbf{85.3}    \\
$\beta$=0.5                       & 96.9          & 85.8          & 91.2            & 87.4            & 90.3             & 95.0          & 77.3          & 87.6            & 77.2            & 84.3             \\
$\beta$=0.0                       & 92.9          & 71.3          & 86.3            & 67.3            & 79.4             & 95.2          & 71.5          & 88.0            & 71.7            & 81.6             \\ \hline
$\gamma$=0.5                      & 91.2          & 75.2          & 71.2            & 61.5            & 74.8             & 93.6          & 63.4          & 82.0            & 74.1            & 78.3             \\
$\gamma$=0.2                      & 96.7          & 83.2          & 90.1            & 92.0            & \textbf{90.5}    & 94.7          & 76.3          & 88.7            & 78.2            & 84.5             \\
$\gamma$=0.0                      & 96.0          & 83.7          & 90.5            & 87.7            & 89.5             & 96.1          & 78.6          & 85.9            & 80.4            & \textbf{85.3}    \\ \hline
\end{tabular}
}
\caption{Experiment results of different reward function settings on Helpsteer2-Pref.}
\label{tab:reward}
\end{table*}

\begin{table}[!h]
\resizebox{0.47\textwidth}{!}{
\begin{tabular}{ccccc}
\hline
\multirow{2}{*}{\textbf{Method}} & \multicolumn{4}{c}{\textbf{RewardBench}}  \\
                        & \textbf{Chat} & \textbf{Safety} & \textbf{Reason} & \textbf{Overall} \\ \hline
Llama-3-8B-Inst         & 64.8 & 76.5   & 63.4   & 68.8    \\ \hline
\multicolumn{5}{l}{\textit{Offline learning on Helpsteer2}} \\ \hline   
iteration 1                 & 81.8 & 88.3   & 77.8   & 83.2    \\
iteration 2                 & 82.5 & 87.2   & 74.8   & 82.5    \\
iteration 3                 & 82.8 & 87.4   & 77.2   & \textbf{83.2}  \\ \hline
\multicolumn{5}{l}{\textit{Offline learning on HH-RLHF}}    \\ \hline
iteration 1                 & 81.2 & 83.2   & 67.0   & 77.3    \\
iteration 2                 & 81.1 & 84.6   & 70.5   & 78.4   \\
iteration 3                 & 78.3 & 83.1   & 72.6   & \textbf{78.9}  \\ \hline
\end{tabular}}
\caption{Experiment results of iterative offline learning.}
\label{tab:iterative-DPO}
\end{table}

We further investigated the impact of different initialization strategies on the subsequent optimization process, as shown in Table \ref{tab:init4RL}. The results indicate that even without thinking initialization, RL-based methods can still achieve substantial improvements, validating their effectiveness. Moreover, initializing the model with a few R1-annotated traces leads to a more structured and effective reasoning pattern, resulting in further enhancements in performance. 



\subsection{Best Practice for Thinking Optimization}

With new RL algorithms continue to emerge in LLM training \cite{zhang2025whathowwherewell}, in this section, we compare different RL strategies for judgment thinking optimization. 

As shown in Table \ref{tab:RL-compare}, for offline learning, removing SFT term will lead to performance degradation, as SFT is conducted on token-level and can provide regularization for better preference optimization.
On the other hand, for online learning, we find that PPO \cite{schulman2017proximalpolicyoptimizationalgorithms} significantly underperforms compared to GRPO and Reinforce++ \cite{hu2025reinforceefficientrlhfalgorithm}. This discrepancy suggests that incorporating a value model for thinking optimization would decrease training stability. We argue that natural language generation (NLG) tasks is distinct from the sequential decision-making tasks in traditional RL. Therefore, the introduction of an additional value model is not only unnecessary but may also hinder training efficiency.


Finally, as shown in Table \ref{tab:iterative-DPO}, offline learning in an iterative manner can achieve further improvement. However, this will result in a more complex and cumbersome training pipeline. To draw a conclusion, the best practice for thinking optimization is GRPO with carefully designed rewards.

\subsection{Reward Design for Online Learning}
\label{sec:indicator-RL}

In this section, we aim to analyze the contribution of different components of the reward function as defined in \ref{equation:reward}. We fixed $\alpha$ as 1.0 and varied the weights $\beta$, and $\gamma$.

As shown in the results in Table \ref{tab:reward}, the formatting reward $r_{\text{format}}$ is crucial for both models, as its removal leads to significant performance degradation. Conversely, the influence of the strength reward $r_{\text{strength}}$ differs between the models. For Qwen-2.5-7B-Inst, removing $r_{\text{strength}}$ results in a performance improvement, while it offers a slight enhancement for the more capable 32B model. We hypothesize that 32B model's stronger inherent abilities allow it to effectively learn the correlation between preference strength and judgment. In contrast, for the weaker 7B model, this additional learning objective may introduce confusion.

\begin{figure}[t]
    \centering
        \includegraphics[width=0.95\linewidth]{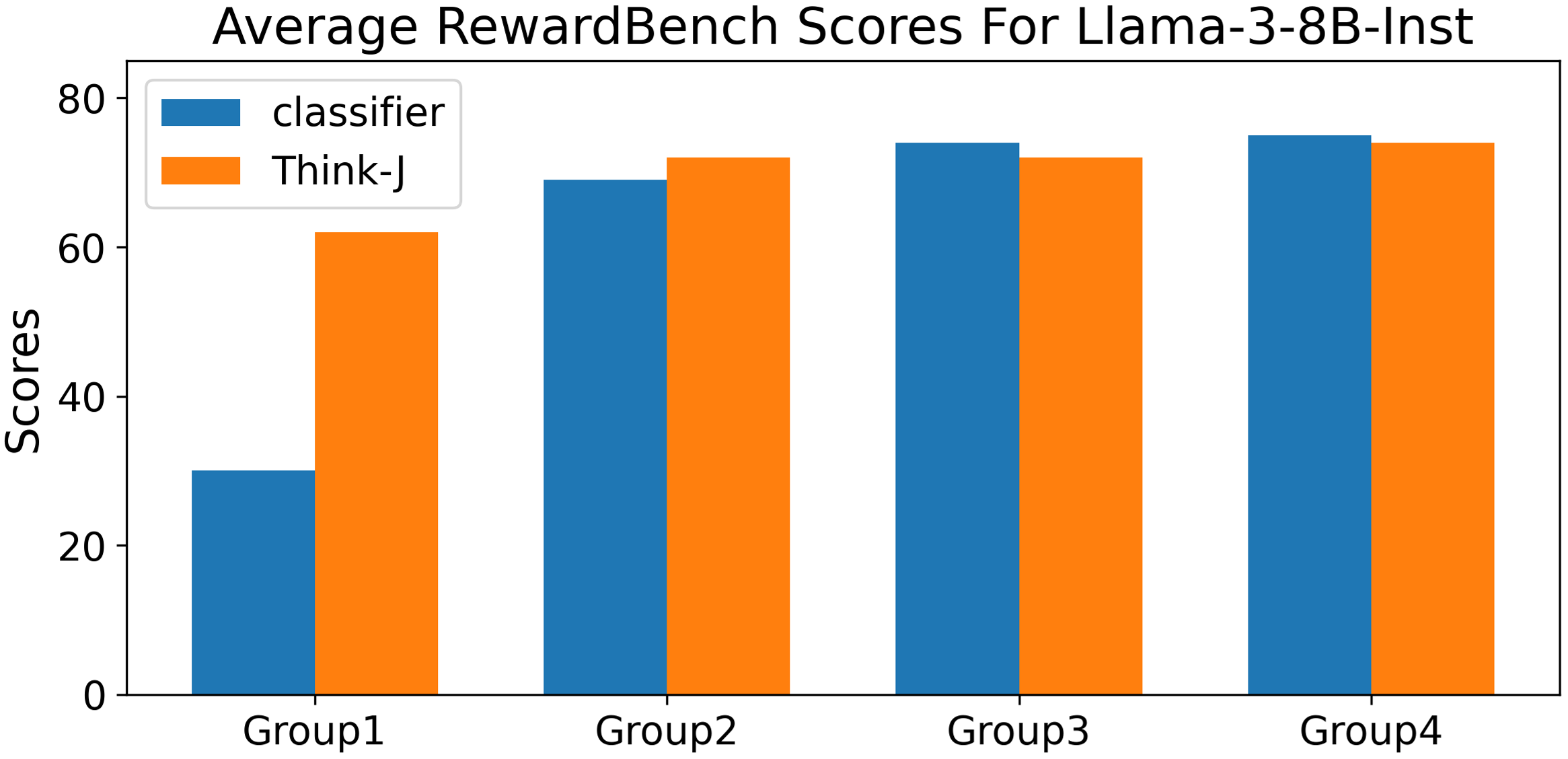}
        \caption{Comparison of different methods on data groups with different quality. Group 1 is with lower quality and Group 4 is with higher quality.}
    \label{figure:bar-chart}
\end{figure}

\subsection{Thinking Makes the Judge More Robust}

In real applications, it is common that there exists noise in the training set, or there is a distributional difference between the training and test sets. In such cases, Think-J demonstrates superior robustness compared with classifiers as a result of its judgment thinking ability.


To verify this, we adopted the approach from \cite{wang-etal-2024-reward-modeling} and divided HH-RLHF into four groups with different data quality\footnote{Data quality is indicated by the difference in scores assigned to response pairs by an external reward model. For more details please refer to the work of \cite{wang-etal-2024-reward-modeling}.}. We then trained judges on the data based on classifier-judge or Think-J. As shown in Figure \ref{figure:bar-chart}, for the two groups with higher data quality, classifier-based judge achieve comparable or even better performance. However, for the groups with lower quality, the accuracy of classifier-based methods drops significantly, even falling below random guessing. In contrast, Think-J maintains relative stability, verifying its robustness to varied data quality\footnote{We also present thinking trace error analysis in Appendix A.6.}.





\section{Conclusion}
In this paper, we propose Think-J to enhance generative LLM-Judges with judgment thinking optimization. Experiment results verify the effectiveness of Think-J compared with both classifier-based and generative LLM judges. With the increasing popularity of RL-based test-time scaling methods, it is crucial to develop a reliable and stable feedback system that aligns well with real-world human preferences. In future, we will continue to explore generative judges for more accurate preference modeling.

\section*{Acknowledgements}
This work was supported in part by the Jiangsu Science and Technology Major Project (BG2024031) and Nanjing University AI \& AI for Science Funding (2024300540).

\bibliography{aaai2026}

@misc{muennighoff2025s1simpletesttimescaling,
      title={s1: Simple test-time scaling}, 
      author={Niklas Muennighoff and Zitong Yang and Weijia Shi and Xiang Lisa Li and Li Fei-Fei and Hannaneh Hajishirzi and Luke Zettlemoyer and Percy Liang and Emmanuel Candès and Tatsunori Hashimoto},
      year={2025},
      eprint={2501.19393},
      archivePrefix={arXiv},
      primaryClass={cs.CL},
      url={https://arxiv.org/abs/2501.19393}, 
}

@misc{ye2025limoreasoning,
      title={LIMO: Less is More for Reasoning}, 
      author={Yixin Ye and Zhen Huang and Yang Xiao and Ethan Chern and Shijie Xia and Pengfei Liu},
      year={2025},
      eprint={2502.03387},
      archivePrefix={arXiv},
      primaryClass={cs.CL},
      url={https://arxiv.org/abs/2502.03387}, 
}

@inproceedings{wang-etal-2024-reward-modeling,
    title = "Reward Modeling Requires Automatic Adjustment Based on Data Quality",
    author = "Wang, Binghai  and
      Zheng, Rui  and
      Chen, Lu  and
      Xi, Zhiheng  and
      Shen, Wei  and
      Zhou, Yuhao  and
      Yan, Dong  and
      Gui, Tao  and
      Zhang, Qi  and
      Huang, Xuanjing",
    editor = "Al-Onaizan, Yaser  and
      Bansal, Mohit  and
      Chen, Yun-Nung",
    booktitle = "Findings of the Association for Computational Linguistics: EMNLP 2024",
    month = nov,
    year = "2024",
    address = "Miami, Florida, USA",
    publisher = "Association for Computational Linguistics",
    url = "https://aclanthology.org/2024.findings-emnlp.234/",
    doi = "10.18653/v1/2024.findings-emnlp.234",
    pages = "4041--4064",
}

@misc{alpaca_eval,
  author = {Xuechen Li and Tianyi Zhang and Yann Dubois and Rohan Taori and Ishaan Gulrajani and Carlos Guestrin and Percy Liang and Tatsunori B. Hashimoto },
  title = {AlpacaEval: An Automatic Evaluator of Instruction-following Models},
  year = {2023},
  publisher = {GitHub},
  journal = {GitHub repository},
  howpublished = {\url{https://github.com/tatsu-lab/alpaca_eval}}
}

@article{zheng2023judging,
  title={Judging LLM-as-a-judge with MT-Bench and Chatbot Arena},
  author={Zheng, Lianmin and Chiang, Wei-Lin and Sheng, Ying and Zhuang, Siyuan and Wu, Zhanghao and Zhuang, Yonghao and Lin, Zi and Li, Zhuohan and Li, Dacheng and Xing, Eric and others},
  journal={arXiv preprint arXiv:2306.05685},
  year={2023}
}

@article{achiam2023gpt,
  title={Gpt-4 technical report},
  author={Achiam, Josh and Adler, Steven and Agarwal, Sandhini and Ahmad, Lama and Akkaya, Ilge and Aleman, Florencia Leoni and Almeida, Diogo and Altenschmidt, Janko and Altman, Sam and Anadkat, Shyamal and others},
  journal={arXiv preprint arXiv:2303.08774},
  year={2023}
}

@article{kim2023prometheus,
  title={Prometheus: Inducing Fine-grained Evaluation Capability in Language Models},
  author={Kim, Seungone and Shin, Jamin and Cho, Yejin and Jang, Joel and Longpre, Shayne and Lee, Hwaran and Yun, Sangdoo and Shin, Seongjin and Kim, Sungdong and Thorne, James and others},
  journal={arXiv preprint arXiv:2310.08491},
  year={2023}
}

@article{li2023generative,
  title={Generative Judge for Evaluating Alignment},
  author={Li, Junlong and Sun, Shichao and Yuan, Weizhe and Fan, Run-Ze and Zhao, Hai and Liu, Pengfei},
  journal={arXiv preprint arXiv:2310.05470},
  year={2023}
}

@article{zhu2023judgelm,
  title={Judgelm: Fine-tuned large language models are scalable judges},
  author={Zhu, Lianghui and Wang, Xinggang and Wang, Xinlong},
  journal={arXiv preprint arXiv:2310.17631},
  year={2023}
}

@article{ouyang2022training,
  title={Training language models to follow instructions with human feedback},
  author={Ouyang, Long and Wu, Jeffrey and Jiang, Xu and Almeida, Diogo and Wainwright, Carroll and Mishkin, Pamela and Zhang, Chong and Agarwal, Sandhini and Slama, Katarina and Ray, Alex and others},
  journal={Advances in Neural Information Processing Systems},
  volume={35},
  pages={27730--27744},
  year={2022}
}

@article{zhu2023promptbench2,
  title={PromptBench: A Unified Library for Evaluation of Large Language Models},
  author={Zhu, Kaijie and Zhao, Qinlin and Chen, Hao and Wang, Jindong and Xie, Xing},
  journal={arXiv preprint arXiv:2312.07910},
  year={2023}
}

@inproceedings{ke-etal-2024-critiquellm,
    title = "{C}ritique{LLM}: Towards an Informative Critique Generation Model for Evaluation of Large Language Model Generation",
    author = "Ke, Pei  and
      Wen, Bosi  and
      Feng, Andrew  and
      Liu, Xiao  and
      Lei, Xuanyu  and
      Cheng, Jiale  and
      Wang, Shengyuan  and
      Zeng, Aohan  and
      Dong, Yuxiao  and
      Wang, Hongning  and
      Tang, Jie  and
      Huang, Minlie",
    editor = "Ku, Lun-Wei  and
      Martins, Andre  and
      Srikumar, Vivek",
    booktitle = "Proceedings of the 62nd Annual Meeting of the Association for Computational Linguistics (Volume 1: Long Papers)",
    month = aug,
    year = "2024",
    address = "Bangkok, Thailand",
    publisher = "Association for Computational Linguistics",
    url = "https://aclanthology.org/2024.acl-long.704",
    pages = "13034--13054",
}

@misc{huang2024empiricalstudyllmasajudgellm,
      title={An Empirical Study of LLM-as-a-Judge for LLM Evaluation: Fine-tuned Judge Model is not a General Substitute for GPT-4}, 
      author={Hui Huang and Yingqi Qu and Xingyuan Bu and Hongli Zhou and Jing Liu and Muyun Yang and Bing Xu and Tiejun Zhao},
      year={2024},
      eprint={2403.02839},
      archivePrefix={arXiv},
      primaryClass={cs.CL},
      url={https://arxiv.org/abs/2403.02839}, 
}

@misc{lambert2024rewardbench,
      title={RewardBench: Evaluating Reward Models for Language Modeling}, 
      author={Nathan Lambert and Valentina Pyatkin and Jacob Morrison and LJ Miranda and Bill Yuchen Lin and Khyathi Chandu and Nouha Dziri and Sachin Kumar and Tom Zick and Yejin Choi and Noah A. Smith and Hannaneh Hajishirzi},
      year={2024},
      eprint={2403.13787},
      archivePrefix={arXiv},
      primaryClass={cs.LG}
}

@article{liu2024skywork,
  title={Skywork-reward: Bag of tricks for reward modeling in llms},
  author={Liu, Chris Yuhao and Zeng, Liang and Liu, Jiacai and Yan, Rui and He, Jujie and Wang, Chaojie and Yan, Shuicheng and Liu, Yang and Zhou, Yahui},
  journal={arXiv preprint arXiv:2410.18451},
  year={2024}
}

@misc{ye2024improvingrewardmodelssynthetic,
      title={Improving Reward Models with Synthetic Critiques}, 
      author={Zihuiwen Ye and Fraser Greenlee-Scott and Max Bartolo and Phil Blunsom and Jon Ander Campos and Matthias Gallé},
      year={2024},
      eprint={2405.20850},
      archivePrefix={arXiv},
      primaryClass={cs.CL},
      url={https://arxiv.org/abs/2405.20850}, 
}

@misc{wang2024secretsrlhflargelanguage,
      title={Secrets of RLHF in Large Language Models Part II: Reward Modeling}, 
      author={Binghai Wang and Rui Zheng and Lu Chen and Yan Liu and Shihan Dou and Caishuang Huang and Wei Shen and Senjie Jin and Enyu Zhou and Chenyu Shi and Songyang Gao and Nuo Xu and Yuhao Zhou and Xiaoran Fan and Zhiheng Xi and Jun Zhao and Xiao Wang and Tao Ji and Hang Yan and Lixing Shen and Zhan Chen and Tao Gui and Qi Zhang and Xipeng Qiu and Xuanjing Huang and Zuxuan Wu and Yu-Gang Jiang},
      year={2024},
      eprint={2401.06080},
      archivePrefix={arXiv},
      primaryClass={cs.AI},
      url={https://arxiv.org/abs/2401.06080}, 
}

@inproceedings{papineni-etal-2002-bleu,
    title = "{B}leu: a Method for Automatic Evaluation of Machine Translation",
    author = "Papineni, Kishore  and
      Roukos, Salim  and
      Ward, Todd  and
      Zhu, Wei-Jing",
    editor = "Isabelle, Pierre  and
      Charniak, Eugene  and
      Lin, Dekang",
    booktitle = "Proceedings of the 40th Annual Meeting of the Association for Computational Linguistics",
    month = jul,
    year = "2002",
    address = "Philadelphia, Pennsylvania, USA",
    publisher = "Association for Computational Linguistics",
    url = "https://aclanthology.org/P02-1040/",
    doi = "10.3115/1073083.1073135",
    pages = "311--318"
}

@article{jaech2024openai,
  title={Openai o1 system card},
  author={Jaech, Aaron and Kalai, Adam and Lerer, Adam and Richardson, Adam and El-Kishky, Ahmed and Low, Aiden and Helyar, Alec and Madry, Aleksander and Beutel, Alex and Carney, Alex and others},
  journal={arXiv preprint arXiv:2412.16720},
  year={2024}
}

@article{guo2025deepseek,
  title={Deepseek-r1: Incentivizing reasoning capability in llms via reinforcement learning},
  author={Guo, Daya and Yang, Dejian and Zhang, Haowei and Song, Junxiao and Zhang, Ruoyu and Xu, Runxin and Zhu, Qihao and Ma, Shirong and Wang, Peiyi and Bi, Xiao and others},
  journal={arXiv preprint arXiv:2501.12948},
  year={2025}
}

@misc{deepseek-math,
      title={DeepSeekMath: Pushing the Limits of Mathematical Reasoning in Open Language Models}, 
      author={Zhihong Shao and Peiyi Wang and Qihao Zhu and Runxin Xu and Junxiao Song and Xiao Bi and Haowei Zhang and Mingchuan Zhang and Y. K. Li and Y. Wu and Daya Guo},
      year={2024},
      eprint={2402.03300},
      archivePrefix={arXiv},
      primaryClass={cs.CL},
      url={https://arxiv.org/abs/2402.03300}, 
}

@inproceedings{
    rafailov2023direct,
    title={Direct Preference Optimization: Your Language Model is Secretly a Reward Model},
    author={Rafael Rafailov and Archit Sharma and Eric Mitchell and Christopher D Manning and Stefano Ermon and Chelsea Finn},
    booktitle={Thirty-seventh Conference on Neural Information Processing Systems},
    year={2023},
    url={https://openreview.net/forum?id=HPuSIXJaa9}
}

@misc{qwen2025qwen25technicalreport,
      title={Qwen2.5 Technical Report}, 
      author={Qwen and : and An Yang and Baosong Yang and Beichen Zhang and Binyuan Hui and Bo Zheng and Bowen Yu and Chengyuan Li and Dayiheng Liu and Fei Huang and Haoran Wei and Huan Lin and Jian Yang and Jianhong Tu and Jianwei Zhang and Jianxin Yang and Jiaxi Yang and Jingren Zhou and Junyang Lin and Kai Dang and Keming Lu and Keqin Bao and Kexin Yang and Le Yu and Mei Li and Mingfeng Xue and Pei Zhang and Qin Zhu and Rui Men and Runji Lin and Tianhao Li and Tianyi Tang and Tingyu Xia and Xingzhang Ren and Xuancheng Ren and Yang Fan and Yang Su and Yichang Zhang and Yu Wan and Yuqiong Liu and Zeyu Cui and Zhenru Zhang and Zihan Qiu},
      year={2025},
      eprint={2412.15115},
      archivePrefix={arXiv},
      primaryClass={cs.CL},
      url={https://arxiv.org/abs/2412.15115}, 
}

@article{grattafiori2024llama,
      title={The llama 3 herd of models},
      author={Grattafiori, Aaron and Dubey, Abhimanyu and Jauhri, Abhinav and Pandey, Abhinav and Kadian, Abhishek and Al-Dahle, Ahmad and Letman, Aiesha and Mathur, Akhil and Schelten, Alan and Vaughan, Alex and others},
      journal={arXiv e-prints},
      pages={arXiv--2407},
      year={2024}
}

@inproceedings{
    liu2025rmbench,
    title={{RM}-Bench: Benchmarking Reward Models of Language Models with Subtlety and Style},
    author={Yantao Liu and Zijun Yao and Rui Min and Yixin Cao and Lei Hou and Juanzi Li},
    booktitle={The Thirteenth International Conference on Learning Representations},
    year={2025},
    url={https://openreview.net/forum?id=QEHrmQPBdd}
}

@inproceedings{
    liu2024statistical,
    title={Statistical Rejection Sampling Improves Preference Optimization},
    author={Tianqi Liu and Yao Zhao and Rishabh Joshi and Misha Khalman and Mohammad Saleh and Peter J Liu and Jialu Liu},
    booktitle={The Twelfth International Conference on Learning Representations},
    year={2024},
    url={https://openreview.net/forum?id=xbjSwwrQOe}
}

@misc{zhang2025whathowwherewell,
      title={What, How, Where, and How Well? A Survey on Test-Time Scaling in Large Language Models}, 
      author={Qiyuan Zhang and Fuyuan Lyu and Zexu Sun and Lei Wang and Weixu Zhang and Zhihan Guo and Yufei Wang and Niklas Muennighoff and Irwin King and Xue Liu and Chen Ma},
      year={2025},
      eprint={2503.24235},
      archivePrefix={arXiv},
      primaryClass={cs.CL},
      url={https://arxiv.org/abs/2503.24235}, 
}

@misc{schulman2017proximalpolicyoptimizationalgorithms,
      title={Proximal Policy Optimization Algorithms}, 
      author={John Schulman and Filip Wolski and Prafulla Dhariwal and Alec Radford and Oleg Klimov},
      year={2017},
      eprint={1707.06347},
      archivePrefix={arXiv},
      primaryClass={cs.LG},
      url={https://arxiv.org/abs/1707.06347}, 
}

@misc{hu2025reinforceefficientrlhfalgorithm,
      title={REINFORCE++: An Efficient RLHF Algorithm with Robustness to Both Prompt and Reward Models}, 
      author={Jian Hu and Jason Klein Liu and Wei Shen},
      year={2025},
      eprint={2501.03262},
      archivePrefix={arXiv},
      primaryClass={cs.CL},
      url={https://arxiv.org/abs/2501.03262}, 
}

@misc{wang2024selftaughtevaluators,
      title={Self-Taught Evaluators}, 
      author={Tianlu Wang and Ilia Kulikov and Olga Golovneva and Ping Yu and Weizhe Yuan and Jane Dwivedi-Yu and Richard Yuanzhe Pang and Maryam Fazel-Zarandi and Jason Weston and Xian Li},
      year={2024},
      eprint={2408.02666},
      archivePrefix={arXiv},
      primaryClass={cs.CL},
      url={https://arxiv.org/abs/2408.02666}, 
}

@inproceedings{park-etal-2024-offsetbias,
    title = "{O}ffset{B}ias: Leveraging Debiased Data for Tuning Evaluators",
    author = "Park, Junsoo  and
      Jwa, Seungyeon  and
      Meiying, Ren  and
      Kim, Daeyoung  and
      Choi, Sanghyuk",
    editor = "Al-Onaizan, Yaser  and
      Bansal, Mohit  and
      Chen, Yun-Nung",
    booktitle = "Findings of the Association for Computational Linguistics: EMNLP 2024",
    month = nov,
    year = "2024",
    address = "Miami, Florida, USA",
    publisher = "Association for Computational Linguistics",
    url = "https://aclanthology.org/2024.findings-emnlp.57/",
    doi = "10.18653/v1/2024.findings-emnlp.57",
    pages = "1043--1067",
}

@article{gu2024survey,
  title={A survey on llm-as-a-judge},
  author={Gu, Jiawei and Jiang, Xuhui and Shi, Zhichao and Tan, Hexiang and Zhai, Xuehao and Xu, Chengjin and Li, Wei and Shen, Yinghan and Ma, Shengjie and Liu, Honghao and others},
  journal={arXiv preprint arXiv:2411.15594},
  year={2024}
}

@misc{cao2024compass,
      title={CompassJudger-1: All-in-one Judge Model Helps Model Evaluation and Evolution}, 
      author={Maosong Cao and Alexander Lam and Haodong Duan and Hongwei Liu and Songyang Zhang and Kai Chen},
      year={2024},
      eprint={2410.16256},
      archivePrefix={arXiv},
      primaryClass={cs.CL},
      url={https://arxiv.org/abs/2410.16256}, 
}

@misc{wang2024helpsteer2,
      title={HelpSteer2-Preference: Complementing Ratings with Preferences}, 
      author={Zhilin Wang and Alexander Bukharin and Olivier Delalleau and Daniel Egert and Gerald Shen and Jiaqi Zeng and Oleksii Kuchaiev and Yi Dong},
      year={2024},
      eprint={2410.01257},
      archivePrefix={arXiv},
      primaryClass={cs.LG},
      url={https://arxiv.org/abs/2410.01257}, 
}

@misc{wang2025helpsteer3,
      title={Dedicated Feedback and Edit Models Empower Inference-Time Scaling for Open-Ended General-Domain Tasks},
      author={Zhilin Wang and Jiaqi Zeng and Olivier Delalleau and Daniel Egert and Ellie Evans and Hoo-Chang Shin and Felipe Soares and Yi Dong and Oleksii Kuchaiev},
      year={2025},
      eprint={2503.04378},
      archivePrefix={arXiv},
      primaryClass={cs.CL},
      url={https://arxiv.org/abs/2503.04378},
}

@misc{chang2023surveyevaluationlargelanguage,
      title={A Survey on Evaluation of Large Language Models}, 
      author={Yupeng Chang and Xu Wang and Jindong Wang and Yuan Wu and Linyi Yang and Kaijie Zhu and Hao Chen and Xiaoyuan Yi and Cunxiang Wang and Yidong Wang and Wei Ye and Yue Zhang and Yi Chang and Philip S. Yu and Qiang Yang and Xing Xie},
      year={2023},
      eprint={2307.03109},
      archivePrefix={arXiv},
      primaryClass={cs.CL},
      url={https://arxiv.org/abs/2307.03109}, 
}

@misc{liu2025inferencetimescalinggeneralistreward,
      title={Inference-Time Scaling for Generalist Reward Modeling}, 
      author={Zijun Liu and Peiyi Wang and Runxin Xu and Shirong Ma and Chong Ruan and Peng Li and Yang Liu and Yu Wu},
      year={2025},
      eprint={2504.02495},
      archivePrefix={arXiv},
      primaryClass={cs.CL},
      url={https://arxiv.org/abs/2504.02495}, 
}

@article{sheng2024hybridflow,
  title={Hybridflow: A flexible and efficient rlhf framework},
  author={Sheng, Guangming and Zhang, Chi and Ye, Zilingfeng and Wu, Xibin and Zhang, Wang and Zhang, Ru and Peng, Yanghua and Lin, Haibin and Wu, Chuan},
  journal={arXiv preprint arXiv:2409.19256},
  year={2024}
}

@inproceedings{zheng2024llamafactory,
  title={LlamaFactory: Unified Efficient Fine-Tuning of 100+ Language Models},
  author={Zheng, Yaowei and Zhang, Richong and Zhang, Junhao and YeYanhan, YeYanhan and Luo, Zheyan},
  booktitle={Proceedings of the 62nd Annual Meeting of the Association for Computational Linguistics (Volume 3: System Demonstrations)},
  pages={400--410},
  year={2024}
}

@misc{whitehouse2025j1incentivizingthinkingllmasajudge,
      title={J1: Incentivizing Thinking in LLM-as-a-Judge via Reinforcement Learning}, 
      author={Chenxi Whitehouse and Tianlu Wang and Ping Yu and Xian Li and Jason Weston and Ilia Kulikov and Swarnadeep Saha},
      year={2025},
      eprint={2505.10320},
      archivePrefix={arXiv},
      primaryClass={cs.CL},
      url={https://arxiv.org/abs/2505.10320}, 
}

@misc{chen2025rmr1rewardmodelingreasoning,
      title={RM-R1: Reward Modeling as Reasoning}, 
      author={Xiusi Chen and Gaotang Li and Ziqi Wang and Bowen Jin and Cheng Qian and Yu Wang and Hongru Wang and Yu Zhang and Denghui Zhang and Tong Zhang and Hanghang Tong and Heng Ji},
      year={2025},
      eprint={2505.02387},
      archivePrefix={arXiv},
      primaryClass={cs.CL},
      url={https://arxiv.org/abs/2505.02387}, 
}

@misc{chen2025judgelrmlargereasoningmodels,
      title={JudgeLRM: Large Reasoning Models as a Judge}, 
      author={Nuo Chen and Zhiyuan Hu and Qingyun Zou and Jiaying Wu and Qian Wang and Bryan Hooi and Bingsheng He},
      year={2025},
      eprint={2504.00050},
      archivePrefix={arXiv},
      primaryClass={cs.CL},
      url={https://arxiv.org/abs/2504.00050}, 
}

@misc{wang2025unifiedmultimodalchainofthoughtreward,
      title={Unified Multimodal Chain-of-Thought Reward Model through Reinforcement Fine-Tuning}, 
      author={Yibin Wang and Zhimin Li and Yuhang Zang and Chunyu Wang and Qinglin Lu and Cheng Jin and Jiaqi Wang},
      year={2025},
      eprint={2505.03318},
      archivePrefix={arXiv},
      primaryClass={cs.CV},
      url={https://arxiv.org/abs/2505.03318}, 
}

@misc{sun2025rethinkingbradleyterrymodelspreferencebased,
      title={Rethinking Bradley-Terry Models in Preference-Based Reward Modeling: Foundations, Theory, and Alternatives}, 
      author={Hao Sun and Yunyi Shen and Jean-Francois Ton},
      year={2025},
      eprint={2411.04991},
      archivePrefix={arXiv},
      primaryClass={cs.AI},
      url={https://arxiv.org/abs/2411.04991}, 
}

@misc{chiang2024chatbotarenaopenplatform,
      title={Chatbot Arena: An Open Platform for Evaluating LLMs by Human Preference}, 
      author={Wei-Lin Chiang and Lianmin Zheng and Ying Sheng and Anastasios Nikolas Angelopoulos and Tianle Li and Dacheng Li and Hao Zhang and Banghua Zhu and Michael Jordan and Joseph E. Gonzalez and Ion Stoica},
      year={2024},
      eprint={2403.04132},
      archivePrefix={arXiv},
      primaryClass={cs.AI},
      url={https://arxiv.org/abs/2403.04132}, 
}

@inproceedings{he2025can,
    title={Can large language models detect errors in long chain-of-thought reasoning?},
    author={He, Yancheng and Li, Shilong and Liu, Jiaheng and Wang, Weixun and Bu, Xingyuan and Zhang, Ge and Peng, Zy and Zhang, Zhaoxiang and Zheng, Zhicheng and Su, Wenbo and others},
    booktitle={Proceedings of the 63rd Annual Meeting of the Association for Computational Linguistics (Volume 1: Long Papers)},
    pages={18468--18489},
    year={2025}
}

@misc{liu2025itrickstrapsdeep,
      title={Part I: Tricks or Traps? A Deep Dive into RL for LLM Reasoning}, 
      author={Zihe Liu and Jiashun Liu and Yancheng He and Weixun Wang and Jiaheng Liu and Ling Pan and Xinyu Hu and Shaopan Xiong and Ju Huang and Jian Hu and Shengyi Huang and Johan Obando-Ceron and Siran Yang and Jiamang Wang and Wenbo Su and Bo Zheng},
      year={2025},
      eprint={2508.08221},
      archivePrefix={arXiv},
      primaryClass={cs.LG},
      url={https://arxiv.org/abs/2508.08221}, 
}

@inproceedings{liu2025air,
  title={Air: Complex instruction generation via automatic iterative refinement},
  author={Liu, Wei and He, Yancheng and Li, Yu and Huang, Hui and Hu, Chengwei and Liu, Jiaheng and Li, Shilong and Su, Wenbo and Zheng, Bo},
  booktitle={Proceedings of the 2025 Conference on Empirical Methods in Natural Language Processing},
  pages={31952--31974},
  year={2025}
}

@inproceedings{huang-etal-2025-empirical,
    title = "An Empirical Study of {LLM}-as-a-Judge for {LLM} Evaluation: Fine-tuned Judge Model is not a General Substitute for {GPT}-4",
    author = "Huang, Hui  and
      Bu, Xingyuan  and
      Zhou, Hongli  and
      Qu, Yingqi  and
      Liu, Jing  and
      Yang, Muyun  and
      Xu, Bing  and
      Zhao, Tiejun",
    editor = "Che, Wanxiang  and
      Nabende, Joyce  and
      Shutova, Ekaterina  and
      Pilehvar, Mohammad Taher",
    booktitle = "Findings of the Association for Computational Linguistics: ACL 2025",
    month = jul,
    year = "2025",
    address = "Vienna, Austria",
    publisher = "Association for Computational Linguistics",
    url = "https://aclanthology.org/2025.findings-acl.306/",
    doi = "10.18653/v1/2025.findings-acl.306",
    pages = "5880--5895",
    ISBN = "979-8-89176-256-5",
}
\clearpage
\appendix
\onecolumn

\section{Additional Experiments}
\subsection{Margin-based Reward for GRPO}
\label{appendix:margin-reward}

As we have explained in Section \ref{sec:online}, we design the reward function as the combination of three parts: $r_{\text{accuracy}}$, $r_{\text{format}}$, $r_\text{strength}$. Referring to the work of \citeauthor{chen2025judgelrmlargereasoningmodels}, for the purpose of more fine-grained judgment prediction, we also tested the following reward function design:

\begin{align}
    r_\text{margin} &= \begin{cases}
        -||\text{s}_{\text{resp1}} - \text{s}_{\text{resp2}}||, & \text{if } \text{judgement} = \text{label} \nonumber \\
        ||\text{s}_{\text{resp1}} - \text{s}_{\text{resp2}}||,  & \text{if } \text{judgement} \neq \text{label}
    \end{cases}
\end{align}



\noindent
where the judgment prediction should also be formatted as:

\vspace{2mm}
\texttt{<think> \{thinking trace\} </think>}

\texttt{Therefore, the quality scores for Response (a) and Response (b) are [[30]] and [[50]], respectively.}
\vspace{2mm}

\begin{table}[h]
\centering
\resizebox{0.98\textwidth}{!}{
\begin{tabular}{c|ccccc|ccccc}
\hline
\multirow{3}{*}{\textbf{Method}} & \multicolumn{5}{c|}{\textbf{Llama-3-8B-Instruct}}                                    & \multicolumn{5}{c}{\textbf{Qwen-2.5-7B-Instruct}}                                    \\
                                 & \multicolumn{5}{c|}{\textbf{RewardBench}}                                            & \multicolumn{5}{c}{\textbf{RewardBench}}                                             \\
                                 & \textbf{Chat} & \textbf{Hard} & \textbf{Safety} & \textbf{Reason} & \textbf{Average} & \textbf{Chat} & \textbf{Hard} & \textbf{Safety} & \textbf{Reason} & \textbf{Average} \\ \hline
online (GRPO w/ margin)              & 96.5          & 66.8          & 88.7            & 73.2            & 81.3             & 96.1          & 72.3          & 83.4            & 76.1            & 82.0             \\
online (GRPO w/o margin)               & 93.9          & 74.3          & 90.3            & 77.8            & \textbf{84.1}    & 96.1          & 78.6          & 85.9            & 80.4            & \textbf{85.3} \\ \hline
\end{tabular}}
\caption{Experiment results of the impact of $r_{margin}$ on Helpsteer2-Pref.}
\label{tab:r_margin}
\end{table}

This design enables fine-grained absolute response scoring, which is more useful for LLM preference optimization pipelines. However, as shown in Table \ref{tab:r_margin}, the introduction of margin-based reward results in performance degradation. By further inspection, we revealed a consistent issue of reward hacking, where the assigned scores were always driven to extreme values (e.g., 0 or 100) to maximize the reward. Despite experimenting with various formulations of $r_{\text{margin}}$, we were unable to effectively mitigate this problem. We believe that mitigating this reward hacking for $r_{\text{margin}}$ necessitates preference data annotated with absolute scores, which we leave for future investigation.

\subsection{Efficiency of Trace Clipping }

As we explained in Section 3.1 and Appendix B.1, we perform trace clipping for the thinking traces generated by Deepseek-R1 to reduce inference overhead and improve traning efficiency. To verify this, we present the results of thinking trace initialization with and without trace clipping in Table \ref{tab:trace-clipping-training}, which clearly shows that when initialization is performed without trace clipping, the optimization effect deteriorates, particularly for online RL. This finding supports our earlier hypothesis that for general LLM judgment, longer thinking do not necessarily correlate with better accuracy. 

\begin{table}[!h]
\centering
\begin{tabular}{cccccc}
\hline
\multirow{2}{*}{\textbf{Init}} & \multirow{2}{*}{\textbf{RL}} & \multicolumn{4}{c}{\textbf{RewardBench}}                            \\
                               &                              & \textbf{Chat} & \textbf{Safety} & \textbf{Reason} & \textbf{Overall} \\ \hline
\multicolumn{2}{c}{Llama-3-8B-Inst}                           & 64.8          & 76.5            & 63.4            & 68.8            \\ \hline
\multirow{2}{*}{w/o clipping}  & offline                      & 82.6          & 88.1            & 70.2            & 81.4            \\
                               & online                       & 79.6          & 85.1            & 67.5            & 78.7            \\ \hline
\multirow{2}{*}{w/ clipping}   & offline                      & 82.8          & 87.4            & 77.2            & 83.2            \\
                               & online                       & 82.9          & 90.3            & 77.8            & \textbf{84.1}   \\ \hline
\end{tabular}
\caption{Comparison of thinking trace initialization with or without trace clipping.}
\label{tab:trace-clipping-training}
\end{table}

Moreover, beyond these performance gains, trace clipping also significantly reduces inference time and computational requirements. To quantify this, we evaluated two online-optimized Qwen-2.5-7B-Instruct models from Table 6 on a single A100-80G GPU. This evaluation involved bidirectional inference across 5,970 samples from RewardBench, and we measured both total inference time and average thinking trace length (tokenized by the Qwen2.5-7B-Instruct tokenizer). As evident from Table \ref{tab:trace-clipping-inference}, trace clipping drastically reduced inference time, primarily due to the considerable decrease in thinking trace length. 

In conclusion, by fostering a more concise thinking process for judgment thinking initialization, trace-clipping not only accelerates inference but also enhances optimization performance.

\begin{table}[!h]
\centering
\begin{tabular}{cccc}
\hline
\textbf{Init} & \textbf{RL} & \textbf{Inference Time(s)} & \textbf{Avg Trace Length} \\ \hline
w/o clip      & online      & 2947.01                    & 907.06                  \\
w/ clip       & online      & 785.42                     & 231.26                  \\ \hline
\end{tabular}
\caption{Comparison of inference efficiency of trace clipping on Qwen-2.5-7B-Instruct.}
\label{tab:trace-clipping-inference}
\end{table}

\subsection{Critic-model for Preference Pair Construction }

While some of previous construct preference pairs for DPO by sampling multiple responses, In Section 3.2, we primarily employ a critic model for constructing the preference pairs. This is because, after judgment thinking initialization, the model already possesses strong judging capabilities. This means simply sampling multiple times often doesn't guarantee the creation of distinct negative examples for much of the preference data.

To verify this, we compared both the efficiency and the performance of our critic-based method and the sampling-based approach on Llama-3-8B-Instruct. Specifically, for each instruction, we sampled 16 judgments, randomly selecting one correct judgment as the positive sample and one incorrect judgment as the negative sample. If a pair of both correct and incorrect judgments couldn't be found, the sample was dropped.

\begin{table}[!ht]
\centering
\begin{tabular}{cccc}
\hline
\textbf{Method} & \textbf{Original} & \textbf{Sampling-based} & \textbf{Critic-based} \\ \hline
Pair Num        & 9125              & 5111                    & 8942                  \\ \hline
\end{tabular}
\caption{Comparison of preference sample number created by different methods on  Llama-3-8B-Instruct.}
\label{tab:critic-datanum}
\end{table}

\begin{table}[!ht]
\centering
\begin{tabular}{cccccc}
\hline
\multirow{2}{*}{\textbf{Method}} & \multicolumn{5}{c}{\textbf{RewardBench}}                                                \\
                                 & \textbf{Chat} & \textbf{Hard} & \textbf{Safety} & \textbf{Reasoning} & \textbf{Overall} \\ \hline
Baseline                         & 90.4          & 44.7          & 76.5            & 63.4               & 68.6            \\ \hline
Sampling-based DPO               & 92.9          & 75.0          & 86.9            & 70.1               & 81.2            \\
Critic-guided DPO                & 95.0          & 73.3          & 87.4            & 77.2               & \textbf{83.2}   \\ \hline
\end{tabular}
\caption{Comparison of offline learning from sampling-based or critic-guided preference pairs on Llama-3-8B-Instruct.}
\label{tab:sampling-training}
\end{table}

\begin{table}[!ht]
\centering
\begin{tabular}{cccccc}
\hline
\multirow{2}{*}{\textbf{Data}} & \multicolumn{5}{c}{\textbf{RewardBench}}                                             \\
                               & \textbf{Chat} & \textbf{Hard} & \textbf{Safety} & \textbf{Reasoning} & \textbf{Overall} \\ \hline
Hybirated                      & 95.0          & 73.3          & 87.4            & 77.2            & 83.2             \\
Solely-Critic                  & 94.9          & 72.6          & 88.1            & 77.6            & 83.3             \\ \hline
\end{tabular}
\caption{Comparison of training of on data constructed solely from the critic model or in a hybrid way on Llama-3-8B-Instruct.}
\label{tab:critic-hybrid}
\end{table}

As shown in Table \ref{tab:critic-datanum}, for the Helpsteer2-Pref dataset, even with n=16 samples, the sampling-based method failed to generate negative samples for nearly half of the data. This significantly reduces data utilization efficiency. Therefore, our critic-based method substantially outperforms the sampling-based method, as demonstrated in Table \ref{tab:sampling-training}. 

Moreover, as we opt to using the critic model to only generate the contrast samples for the preference pairs, we also verified the effectiveness of training solely on data generated by the critic. As shown in Table \ref{tab:critic-hybrid}, there was no significant difference in performance. Therefore, we decided to use the critic to construct either positive or negative samples for a given response. This approach effectively reduces inference time by half compared to constructing both.

\subsection{Statistical Indicators of Judgment Thinking Optimization}

\begin{figure}[H]
    \centering
    \includegraphics[width=0.7\linewidth]{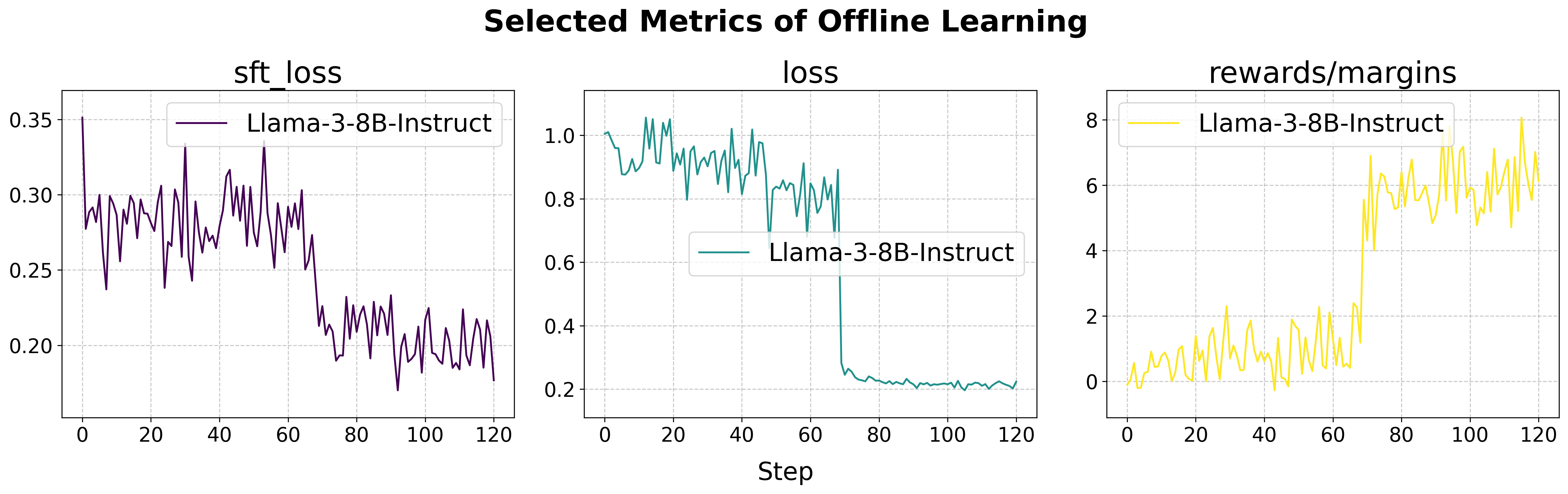}
    \caption{The variation of statistical metrics during offline learning trained on Helpsteer2-Pref.}
    \label{fig: dpo-indicators}
\end{figure}

\begin{figure}[H]
    \centering
    \includegraphics[width=0.7\linewidth]{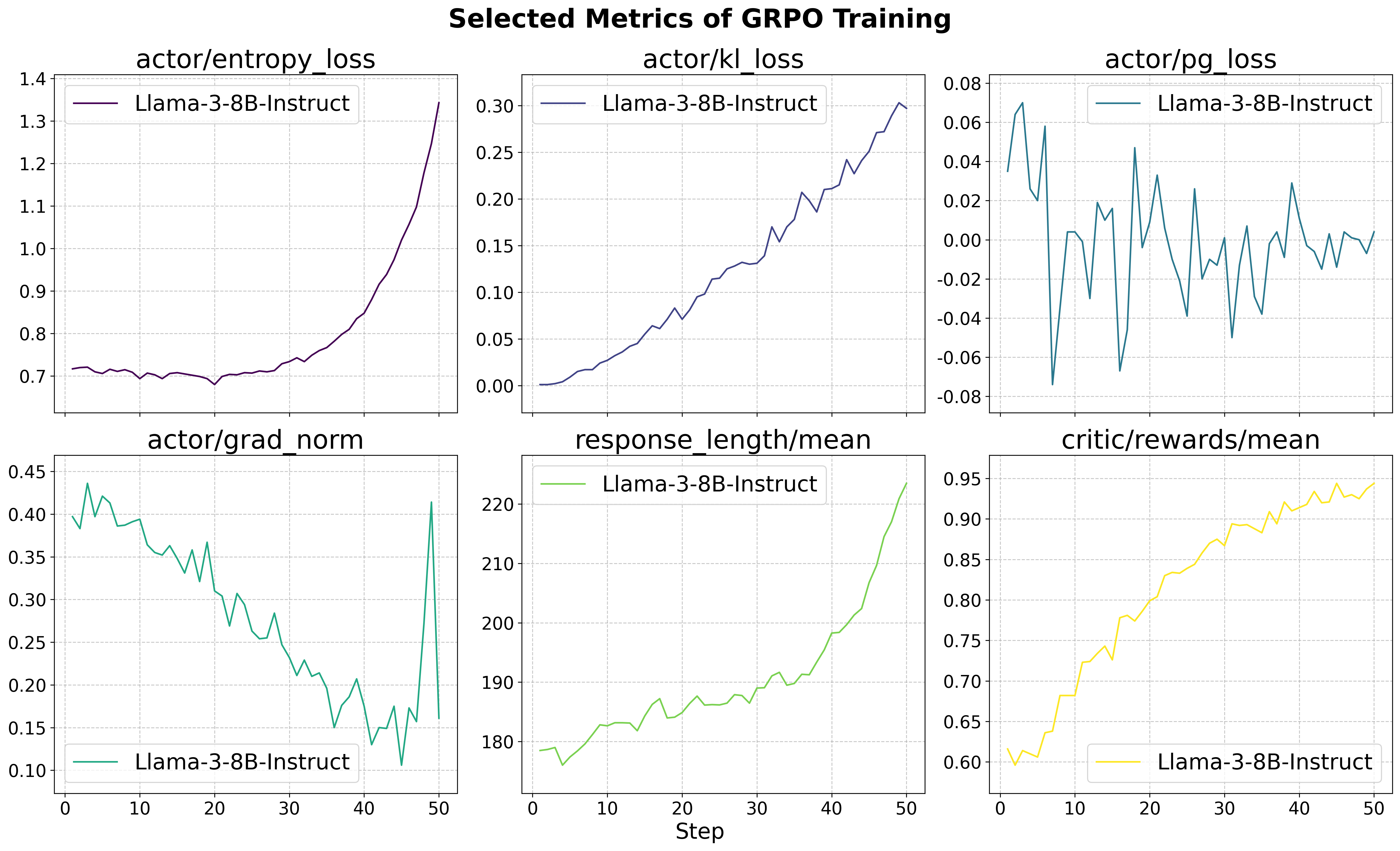}
    \caption{The variation of statistical metrics during GRPO training on Helpsteer2-Pref.}
    \label{fig: grpo-indicators}
\end{figure}

\begin{figure}[H]
    \centering
    \includegraphics[width=0.7\linewidth]{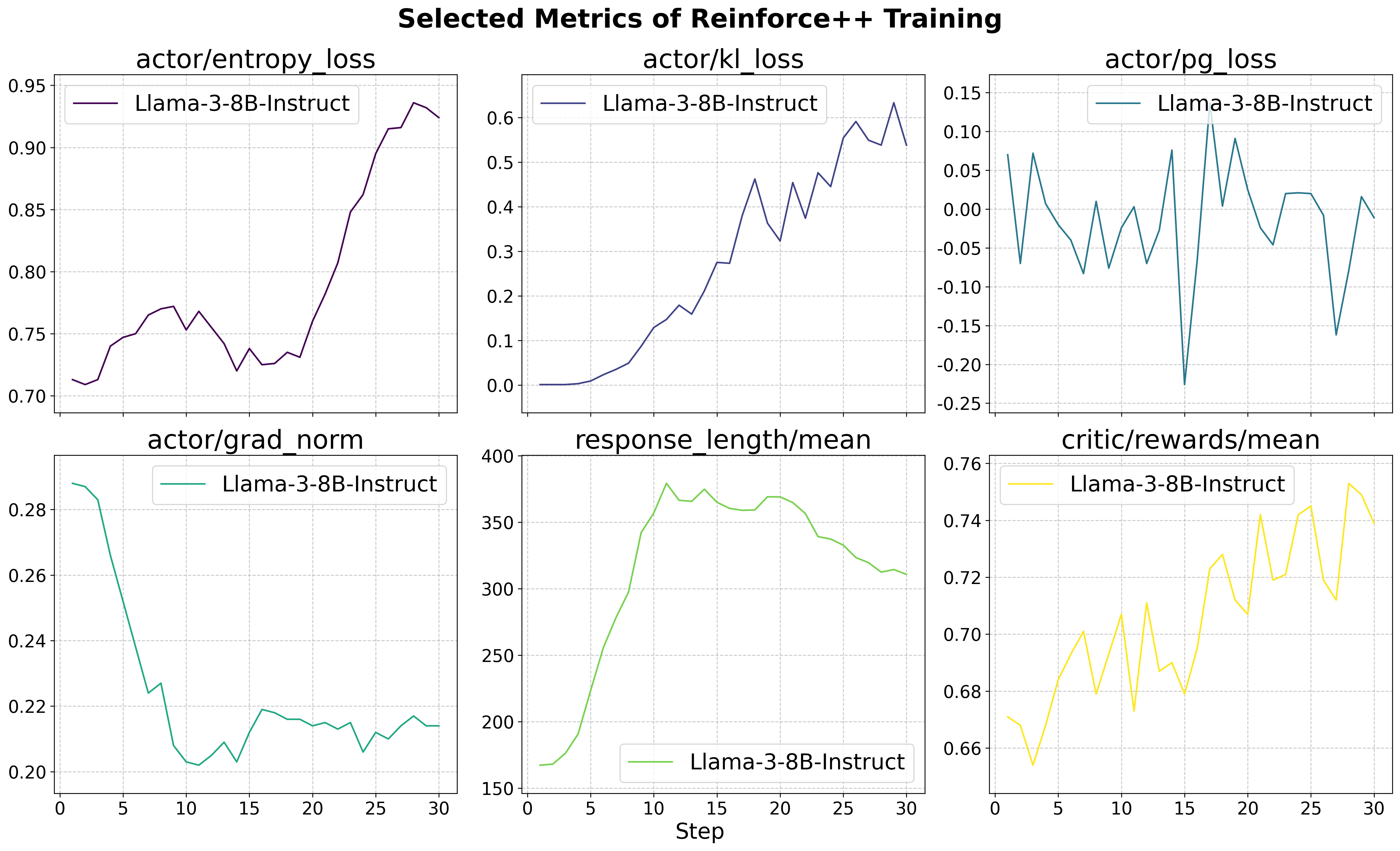}
    \caption{The variation of statistical metrics during Reinforce++ training on Helpsteer2-Pref.}
    \label{fig: rlpp-indicators}
\end{figure}

\begin{figure}[H]
    \centering
    \includegraphics[width=0.92\linewidth]{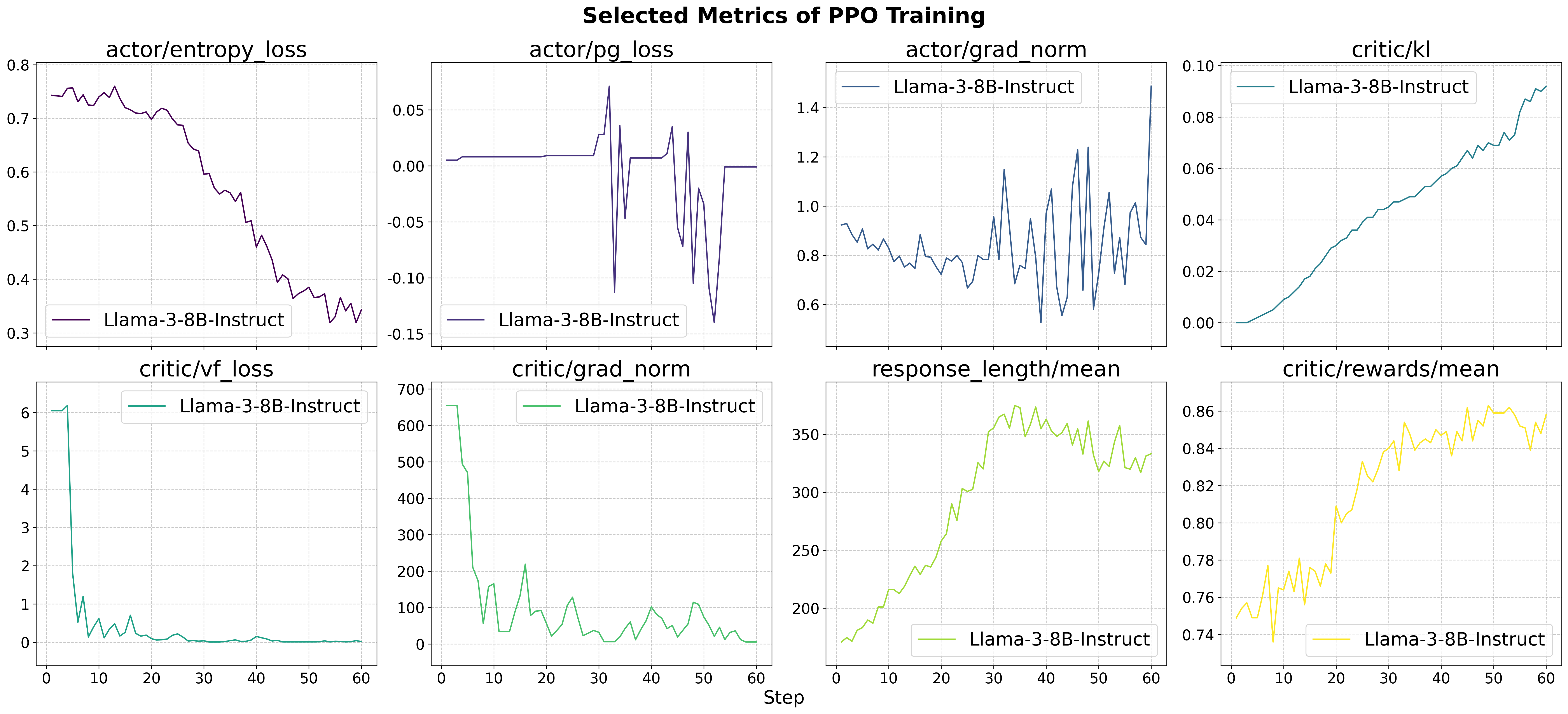}
    \caption{The variation of statistical metrics during PPO training on Helpsteer2-Pref.}
    \label{fig: ppo-indicators}
\end{figure}

In this section, we present the detailed statistical indicators during Judgment Thinking Optimization. The statistical indicators of offline learning, GRPO training, PPO \cite{schulman2017proximalpolicyoptimizationalgorithms} training and Reinforce++ \cite{hu2025reinforceefficientrlhfalgorithm} training are presented in Figure \ref{fig: dpo-indicators}, \ref{fig: grpo-indicators}, \ref{fig: ppo-indicators}, \ref{fig: rlpp-indicators}, respectively. 


As the results indicate, GRPO demonstrates consistent and substantial increases in predicted rewards, a strong signal of learning to generate desirable outputs, and the critic's low and stable value function loss suggests reliable outcome evaluation. While DPO effectively aligns with preferences, and Reinforce++ and PPO show progress in reward acquisition and stability, GRPO's robust reward improvement combined with a well-functioning critic positions it as the most robust algorithm based on these indicators. Notably, GRPO's response length exhibits a consistent gradual increase, similar to prior work of \cite{deepseek-math}, while other methods start to fluctuate after a certain point. This suggests GRPO effectively learns to generate more detailed reasoning, enhancing judgment accuracy.


\subsection{Human Evaluation of Thinking Trace Interpretability}

To quantitatively verify the improvement of thinking trace interpretability introduced by our method, we conducted a human evaluation. We randomly sampled 100 cases from RewardBench and enlisted three graduate students to manually assess the thinking trace produced by three judge models: the original model, the model fine-tuned with Supervised Fine-Tuning (SFT) on positive data only, and our model optimized with online Reinforcement Learning (RL), Think-J. The students were instructed to compare the two judgments given by different models for the same sample and determine which one provided a better thinking. We then recorded the majority vote from the three students.

\begin{table}[!ht]
\centering
\begin{tabular}{cc}
\hline
\textbf{Setting}                       & \textbf{Win Rate} \\ \hline
Think-J v.s. Original Qwen-2.5-7B-Inst & 100\%             \\
Think-J v.s. SFT on Positive Samples   & 67\%              \\ \hline
\end{tabular}
\caption{Comparison of interpretability of thinking traces generated by different judges.}
\label{tab:interpretability}
\end{table}

As the results show in Table \ref{tab:interpretability}, Think-J unquestionably outperforms the judgments generated by the original model. This is expected, as the original model was not optimized for thinking-enhanced judgment, and its interpretations were too simplistic. Furthermore, Think-J also surpasses the judgments from the model SFT on positive samples only, underscoring the necessity of negative samples when optimizing judgment thinking traces.

\subsection{Limited Failure Analysis of Think-J Robustness}

While we generally assume a correct thinking trace would lead to a correct judgment (and vice versa), we acknowledge cases where an incorrect thinking trace might still produce the right outcome. To perform a more in-depth evaluation of the model's underlying thinking process, we leveraged an LLM-as-a-Judge approach based on the data and model in Section 5.4.

\begin{table}[!h]
\centering
\begin{tabular}{ccccc}
\hline
\multirow{2}{*}{\textbf{Model}} & \multicolumn{4}{c}{\textbf{Error Rate}}                              \\
                                & \textbf{Chat} & \textbf{Hard} & \textbf{Safety} & \textbf{Reasoning} \\ \hline
Think-J-Qwen-2.5 trained on Group1                 & 32.5          & 55.0          & 60.0            & 63.1               \\
Think-J-Qwen-2.5 trained on Group4                 & 23.1          & 33.3          & 46.7            & 35.0               \\ \hline
\end{tabular}
\caption{Error rate of thinking traces generated by different Think-J models trained on different qualities groups of data.}
\label{tab:cot-failure}
\end{table}

Specifically, we randomly selected 50 samples from RewardBench and employed GPT-4o to meticulously check for errors in the judgment process (i.e., the thinking trace) for each model. We removed the final binary judgment (i.e., "Response A/B is better") and focused solely on verifying errors within the clipped critique itself, even when the final judgment result was correct. This methodology allowed us to better assess the robustness of Think-J's internal logic.

Unexpectedly, as shown in Table \ref{tab:cot-failure}, our analysis revealed a significant finding: Despite the high accuracy of the model trained on noised Group 1 data, its thinking trace quality appeared to degrade substantially, with over 50\% of the traces containing identifiable errors. We hypothesize that for a binary classification task like judgment, an imperfect thinking trace can still lead to a correct judgment in many cases, provided the critical point with decisive influence is identified. 

\section{Implementation Details}
\label{sec:impl}

\subsection{Thinking Trace Clipping }

As we have explained in Section 3.1, despite the superior reasoning capability of Deepseek-R1, the thinking trace generated are excessively long and could possibly cause difficulty for further optimization and increase computational overhead. Specifically, for our case, we hypothesize that general LLM judgment does not necessarily require a detailed, step-by-step reasoning process. On the contrary, identifying one key factor is sufficient to make correct judgment in most cases. Overly long reasoning traces can cause the model to consider factors that do not affect the preference outcomes, thereby degrading the effectiveness.

\begin{figure}[h]
    \centering
        \includegraphics[width=0.75\linewidth]{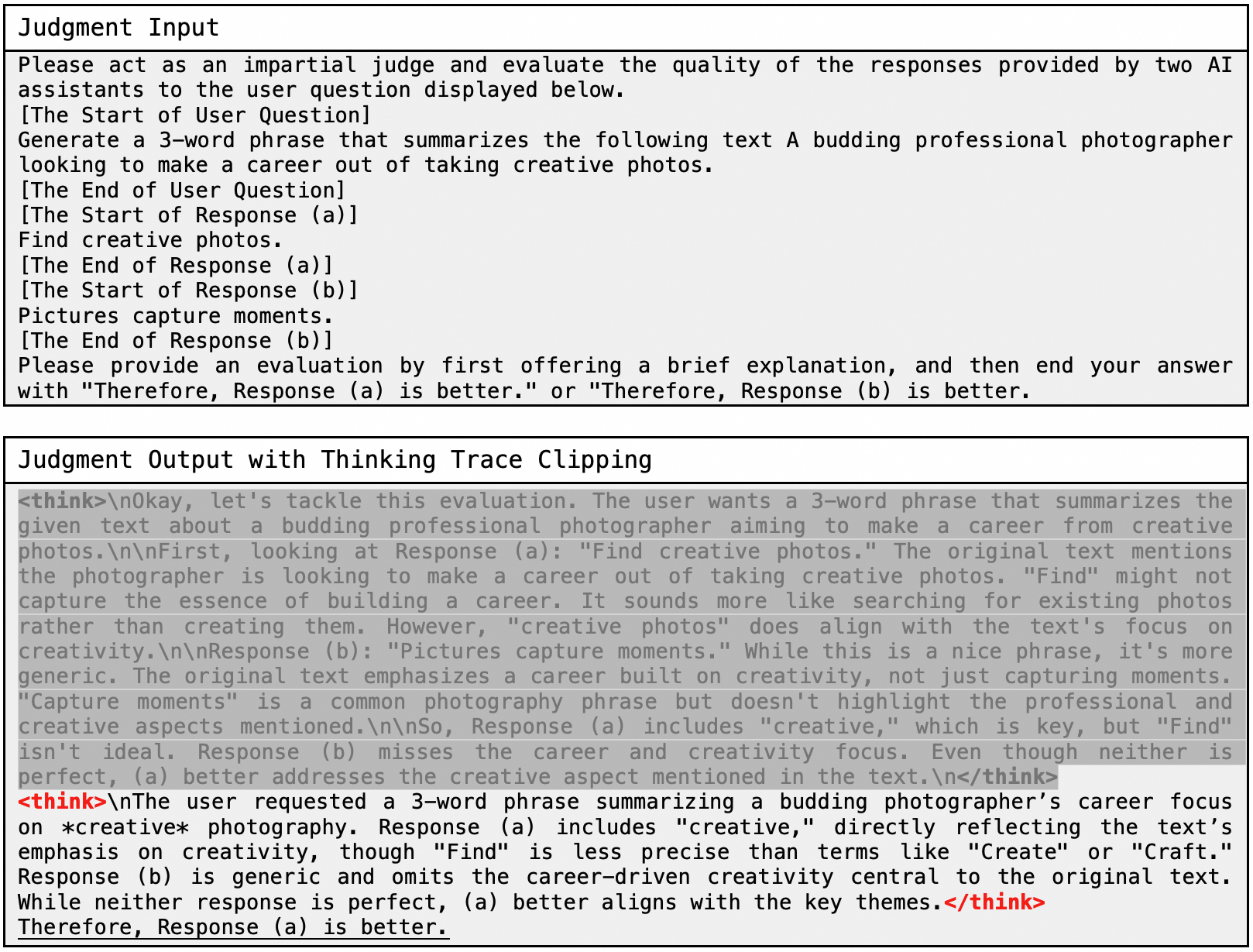}
        \caption{The illustration of Trace Clipping. The first part of thinking trace is removed, while the second part is used as the output thinking trace.}
    \label{figure:reasoning-clipping}
\end{figure}

Therefore, we introduce \textbf{Trace Clipping} to enhance the effectiveness of thinking initialization, as shown in the Example of \ref{figure:reasoning-clipping}. Specifically, we observed that the CoTs generated by R1 are composed of two parts: a lengthy reasoning process, and an explanation which summarizes the reasoning process. Therefore, we remove the first part and use the second part as the output trace. Models trained with the clipped trace will also generate the reasoning trace in a way that directly addresses the key points. This not only facilitates subsequent optimization but also enhances the judgment readability.

The experiment analysis of the effectiveness of thinking trace clipping is presented in Appendix A.2.

\subsection{Judgment Thinking Initialization}
\label{sec:impl1}

As we have explained in Section \ref{sec:think-init}, we apply deduplication to ensure the diversity of LIMJ707. For this purpose, we leverage the algorithm defined in Algorithm \ref{alg:density} to create instruction data pool, where the embeddings of instructions are derived based on the model of instructor-large\footnote{\url{https://huggingface.co/hkunlp/instructor-large}}, and repetitive instructions are removed based on the similarity of instruction semantics.

We leverage the prompt templates in Figure \ref{fig: prompt1} and \ref{fig: prompt3} for constructing thinking trace based on proprietary thinking models. The temperature is set as 0.0 to enable more accurate judgment.

\begin{figure}[htb]
  \vspace{-3mm}
  \centering
  \begin{minipage}{.9\linewidth}
    \begin{algorithm}[H]
      \renewcommand{\algorithmicrequire}{\textbf{Input:}}
      \renewcommand{\algorithmicensure}{\textbf{Output:}}
      \caption{Diversity-verified Sampling}
      \label{alg:1}
      \begin{algorithmic}[1]
        \REQUIRE Instruction Dataset $D$ with $m$ samples, number of samples to select $n$.
        \ENSURE Selected Dataset $D'$ with $n$ samples.
        \STATE Derive the embeddings for each sample in $D$.
        \STATE Random sample one data point $x$ from $D$.
        \STATE Delete $x$ from $D$, add $x$ to $D'$.
        \FOR{$i = 1, 2, ..., n-1$}
          \STATE Calculate the cosine similarity score between $x$ and each sample from $D$.
          \STATE Select the least similar sample $x'$ from $D'$.
          \STATE Let $x = $ $x'$.
          \STATE Delete $x$ from $D$, add $x$ to $D'$.
        \ENDFOR
      \end{algorithmic}
      \label{alg:density}
    \end{algorithm}
  \end{minipage}
  \vspace{-3mm}
  \caption{The sampling algorithm for LIMJ707 used for diversity verification.}
\end{figure}

\subsection{Judgment Thinking Optimization}
\label{sec:impl2}
\begin{figure}[H]
    \centering
    \includegraphics[width=0.8\linewidth]{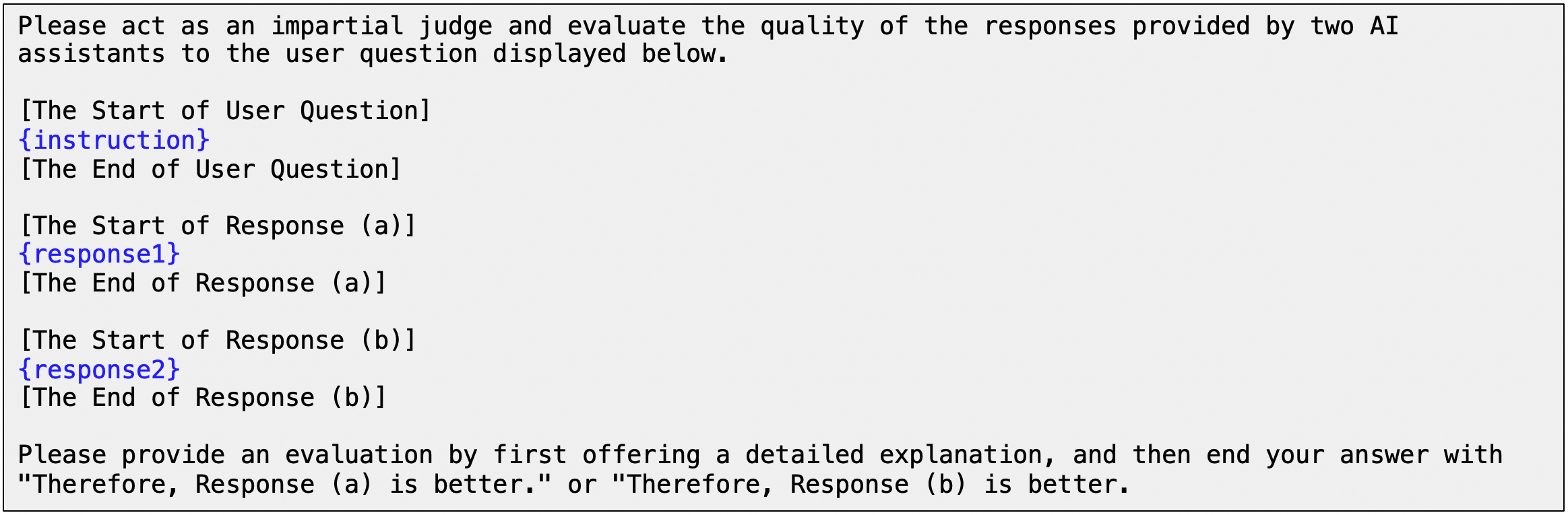}
    \caption{The prompt template used for judgment prediction without $r_{strength}$.}
    \label{fig: prompt1}
\end{figure}

\begin{figure}[H]
    \centering
    \includegraphics[width=0.8\linewidth]{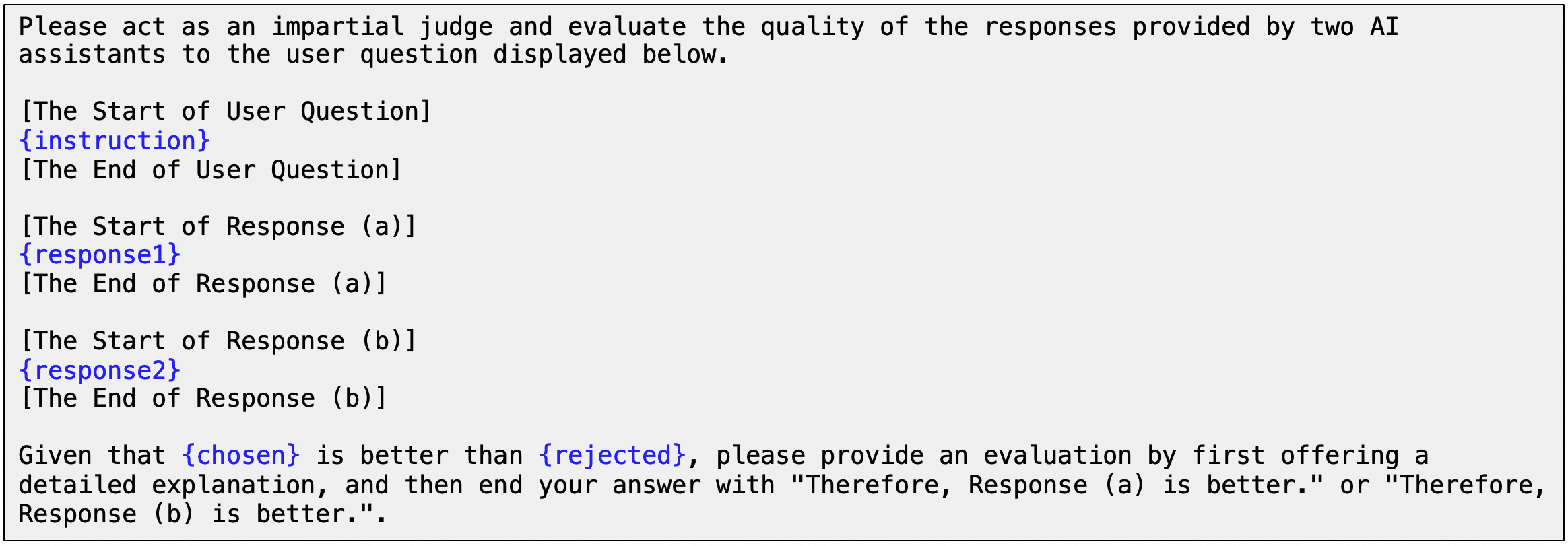}
    \caption{The prompt template used for critique prediction without $r_{strength}$.}
    \label{fig: prompt2}
\end{figure}

\begin{figure}[H]
    \centering
    \includegraphics[width=0.8\linewidth]{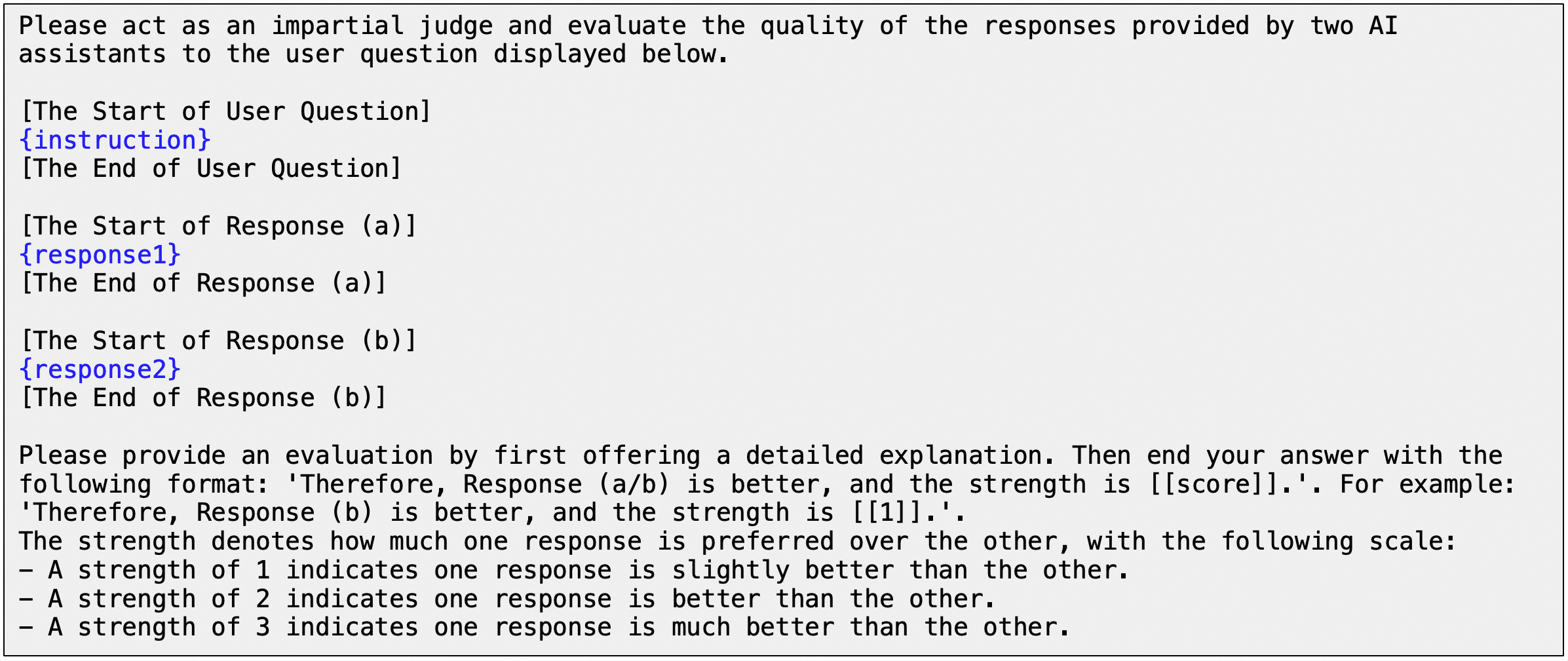}
    \caption{The prompt template used for judgment prediction with $r_{strength}$.}
    \label{fig: prompt3}
\end{figure}

\begin{figure}[H]
    \centering
    \includegraphics[width=0.8\linewidth]{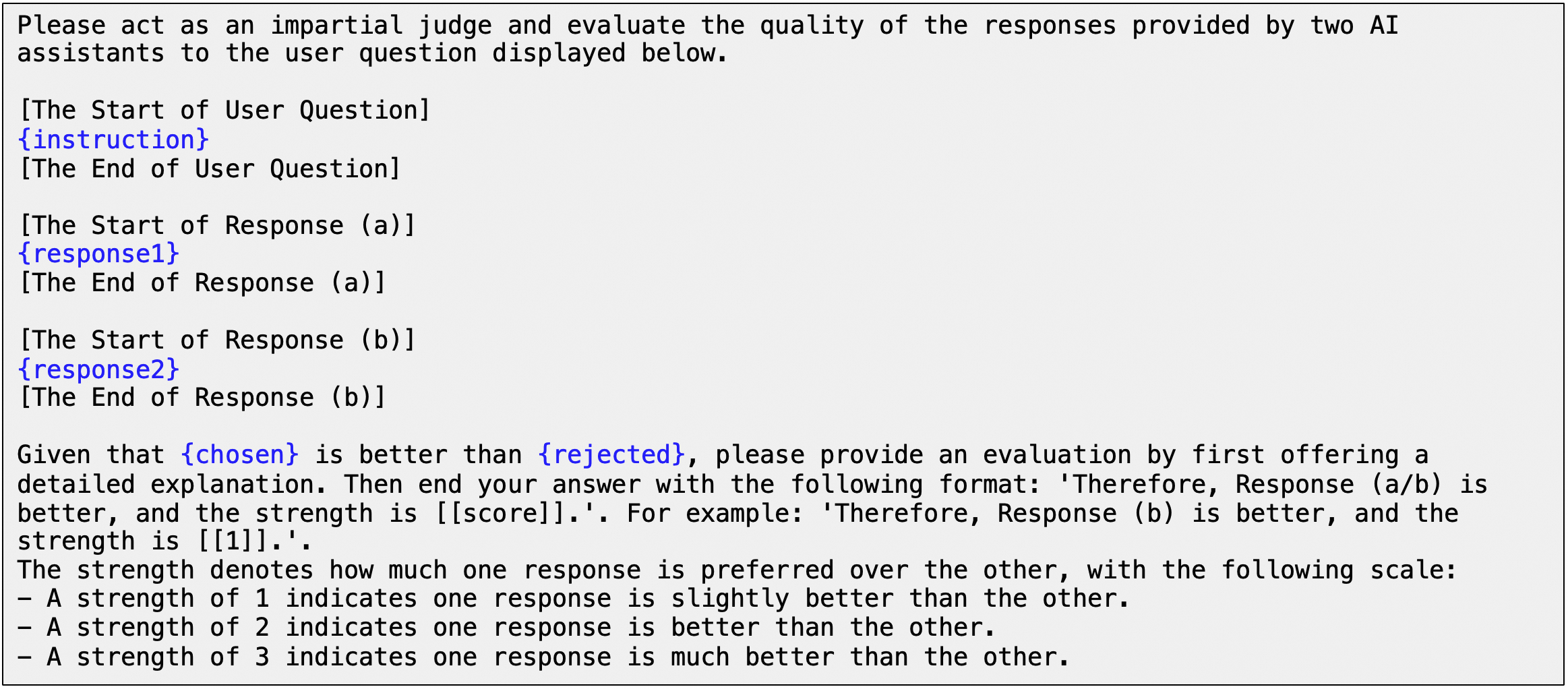}
    \caption{The prompt template used for critique prediction with $r_{strength}$.}
    \label{fig: prompt4}
\end{figure}

For judgment thinking optimization, we leverage the prompts in Figure \ref{fig: prompt1} and \ref{fig: prompt2} for offline learning, and the prompts in Figure \ref{fig: prompt3} and \ref{fig: prompt4} for online learning. The offline learning is implemented based on Llama-Factory \cite{zheng2024llamafactory}, and online learning is implemented based on verl \cite{sheng2024hybridflow}. We present the hyper-parameter settings in Table \ref{tab: params1} and \ref{tab: params2}.

\subsection{Criteria for Determining Preference Strength}

As we have explained in Section 3.2, for Rule-based Online Learning, we propose an $r_{strength}$ for assessing preference strength, which is defined as the degree to which the judge favors one response over another. In this section, we would like to detail how we get the strength annotation for HelpSteer2-Pref and HH-RLHF datasets.

For the HelpSteer2 dataset, we directly adopt the original data annotations, which were derived from professional human annotators and provide three distinct strength levels.

For the HH-RLHF dataset, we derive strength annotations following the method of \cite{wang-etal-2024-reward-modeling}, which involves:

\begin{enumerate}
    \item Training multiple Bradley-Terry classifiers on preference data.
    \item Averaging the scores assigned to each data point by these classifiers.
    \item Categorizing the strength into three tiers (1, 2, 3) based on the score differences between preference pairs.
\end{enumerate}

We opted for a three-level categorization of strength—representing "slightly better," "better," and "much better"—due to its clear interpretability. We found that further granularity did not improve results and, in some cases, led to the model "hacking" to extreme values, as detailed in Appendix B.1.

\begin{table}[h]
\centering
\begin{tabular}{cccc}
\hline
\textbf{hyper-parameter}        & \textbf{SFT} & \textbf{offline} & \textbf{BT classifier} \\ \hline
learning\_rate                  & 5.00E-06     & 1.00E-06         & 2.00E-06               \\
per\_device\_train\_batch\_size & 64           & 64               & 64                     \\
cutoff\_len                     & 4096         & 4096             & 4096                   \\
lr\_scheduler                   & cosine       & cosine           & cosine                 \\
train\_epoch                    & 3            & 2                & 2                      \\
kl\_beta                        & —            & 0.1              & —                      \\
deepspeed\_stage                & zero3        & zero2            & —                      \\
precision                       & bf16         & bf16             & bf16                   \\
flash\_attn                     & fa2          & fa2              & —                      \\ \hline
\end{tabular}
\caption{Hyper-parameter settings for SFT, offline learning and BT classifier.}
\label{tab: params1}
\end{table}

\begin{table}[h]
\centering
\begin{tabular}{cccc}
\hline
\textbf{hyper-parameter}                                            & \textbf{GRPO} & \textbf{PPO} & \textbf{Reinforce++} \\ \hline
data.train\_batch\_size                                             & 1024          & 1024         & 1024                 \\
data.max\_prompt\_length                                            & 1024          & 1024         & 1024                 \\
data.max\_response\_length                                          & 3072          & 3072         & 3072                 \\
actor\_rollout\_ref.actor.optim.lr                                  & 1.00E-06      & 1.00E-06     & 3.00E-06             \\
actor\_rollout\_ref.actor.ppo\_mini\_batch\_size                    & 256           & 256          & 1024                 \\
actor\_rollout\_ref.actor.ppo\_micro\_batch\_size\_per\_gpu         & 16            & 4            & 8                    \\
actor\_rollout\_ref.actor.use\_kl\_loss                             & true          & —            & true                 \\
actor\_rollout\_ref.actor.kl\_loss\_coef                            & 0.001         & —            & 0.001                \\
actor\_rollout\_ref.actor.kl\_loss\_type                            & low\_var\_kl  & —            & mse                  \\
actor\_rollout\_ref.rollout.log\_prob\_micro\_batch\_size\_per\_gpu & 16            & 4            & 8                    \\
actor\_rollout\_ref.rollout.n                                       & 8             & —            & 4                    \\
actor\_rollout\_ref.ref.log\_prob\_micro\_batch\_size\_per\_gpu     & 16            & 4            & 8                    \\
critic.optim.lr                                                     & —             & 1.00E-05     & —                    \\
critic.ppo\_micro\_batch\_size\_per\_gpu                            & —             & 4            & —                    \\
algorithm.kl\_ctrl.kl\_coef                                         & 0.001         & 0.001        & —                    \\
trainer.n\_gpus\_per\_node                                          & 8             & 8            & 8                    \\
trainer.nnodes                                                      & 2             & 2            & 2                    \\
trainer.total\_epochs                                               & 10            & 5            & 5                    \\ \hline
\end{tabular}
\caption{Hyper-parameter settings for online learning algorithms.}
\label{tab: params2}
\end{table}

\end{document}